\documentclass[lettersize,journal]{IEEEtran}
\usepackage{amsmath,amsfonts}
\usepackage{algorithmic}
\usepackage{algorithm}
\usepackage{color}
\usepackage{array}
\usepackage[caption=false,font=normalsize,labelfont=sf,textfont=sf]{subfig}
\usepackage{textcomp}
\usepackage{stfloats}
\usepackage{url}
\usepackage{verbatim}
\usepackage{graphicx}
\usepackage{cite}
\usepackage{hyperref}
\hypersetup{hypertex=true,
            colorlinks=true,
            linkcolor=blue,
            anchorcolor=blue,
            citecolor=blue}
\usepackage{multirow}
\begin{document}

\title{End-to-end Open-vocabulary Video Visual Relationship Detection using Multi-modal Prompting }

\author{Yongqi Wang, Xinxiao Wu,~\IEEEmembership{Member,~IEEE},  Shuo Yang, Jiebo Luo~\IEEEmembership{Fellow,~IEEE}
\thanks{

Yongqi Wang and Xinxiao Wu are with the Beijing Key Laboratory of Intelligent Information Technology, School of Computer Science and Technology, Beijing Institute of Technology, Beijing 100081, China (e-mail: 3120230916@bit.edu.cn, wuxinxiao@bit.edu.cn).

Xinxiao Wu is also with the Guangdong Provincial Laboratory of Machine Perception and Intelligent Computing, Shenzhen MSU-BIT University, Shenzhen 518172, China.

Shuo Yang is with the Guangdong Provincial Laboratory of Machine Perception and Intelligent Computing, Shenzhen MSU-BIT University, Shenzhen 518172, China (e-mail: yangshuo@smbu.edu.cn).

Jiebo Luo is with the Department of Computer Science, University of Rochester, Rochester, NY 14627 USA (e-mail: jluo@cs.rochester.edu).}
}



\maketitle

\begin{abstract}
Open-vocabulary video visual relationship detection aims to expand video visual relationship detection beyond annotated categories by detecting unseen relationships between both seen and unseen objects in videos.
Existing methods usually use trajectory detectors trained on closed datasets to detect object trajectories, and then feed these trajectories into large-scale pre-trained vision-language models 
to achieve open-vocabulary classification. 
Such heavy dependence on the pre-trained trajectory detectors limits their ability to generalize to novel object categories, leading to performance degradation.
To address this challenge, we propose to unify object trajectory detection and relationship classification into an end-to-end open-vocabulary framework.
Under this framework, we propose a relationship-aware open-vocabulary trajectory detector. It primarily consists of a query-based Transformer decoder, where the visual encoder of CLIP is distilled for frame-wise open-vocabulary object detection, and a trajectory associator.
To exploit relationship context during trajectory detection, a relationship query is embedded into the Transformer decoder, and accordingly, an auxiliary relationship loss is designed to enable the decoder to perceive the relationships between objects explicitly. 
Moreover, we propose an open-vocabulary relationship classifier that leverages the rich semantic knowledge of CLIP to discover novel relationships. 
To adapt CLIP well to relationship classification, we design a multi-modal prompting method that employs spatio-temporal visual prompting for visual representation and vision-guided language prompting for language input.
Extensive experiments on two public datasets, VidVRD and VidOR, demonstrate the effectiveness of our framework. 
Our framework is also applied to a more difficult cross-dataset scenario to further demonstrate its generalization ability. 
The code for this paper is available at \url{https://github.com/wangyongqi558/EOV-MMP-VidVRD}.
\end{abstract}


\section{Introduction}
\IEEEPARstart{V}{ideo} Visual Relationship Detection (VidVRD) aims to detect objects and their relationships in videos, typically represented as triplets in the format of $\langle subject, relationship, object \rangle$~\cite{shang2017video}. 
Open-vocabulary Video Visual Relationship Detection (Open-VidVRD) expands VidVRD task by training on base categories of objects and relationships, and testing on both base and novel categories~\cite{gao2023compositional}, 
which has wide applications in real-world scenarios. 

\begin{figure}
\centering
\includegraphics[width=0.95\linewidth]{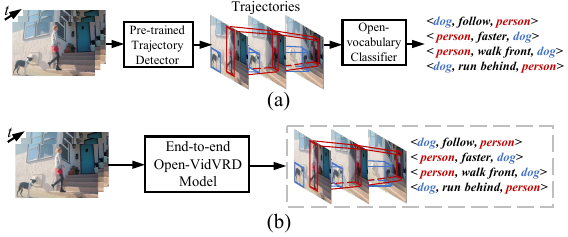}
\caption{(a) Existing Open-VidVRD methods rely on trajectory detectors trained on closed datasets. (b) The proposed end-to-end model performs Open-VidVRD directly on the original videos. }
\label{fig:intro}
\end{figure}

Recent significant progress has been made on open-vocabulary tasks~\cite{zang2022open,li2023ovtrack,zheng2023open} by integrating charming large-scale pre-trained vision-language models~\cite{jia2021scaling, pham2023combined, radford2021learning, li2022blip, li2023blip}. By learning joint vision-language embeddings, these pre-trained models can exploit extensive semantic knowledge of objects, scenes, actions, and interactions~\cite{gao2022open, gu2021open, kuo2022f, ni2022expanding, weng2023open, xu2023side}.
Existing Open-VidVRD methods~\cite{gao2023compositional,yang2024multi} typically employ prompt learning in pre-trained models to facilitate open-vocabulary classification of objects and relationships.
These methods firstly use the trajectory detectors pre-trained on closed datasets to detect object trajectories from videos, and then feed the trajectories into pre-trained models like~\cite{radford2021learning} for open-vocabulary classification of objects and relationships, as illustrated in Figure~\ref{fig:intro}(a).
Such heavy reliance on closed-set trajectory detectors limits their generalization capabilities to unseen object categories. 
Additionally, the domain gap between the training data of trajectory detectors and that of the Open-VidVRD task limits their adaptability to base categories.
As a result, the detected object trajectories are suboptimal, hindering the subsequent relationship classification. 

To address this challenge, we propose a novel end-to-end framework for Open-VidVRD, as illustrated in Figure~\ref{fig:intro}(b). It jointly models object trajectory detection and relationship classification into a unified framework.   
Under this framework, we propose two key components: a relationship-aware open-vocabulary trajectory detector and an open-vocabulary relationship classifier. The trajectory detector primarily consists of a query-based Transformer decoder in which the visual encoder of CLIP is distilled for frame-wise open-vocabulary object detection, and a trajectory associator for generating trajectories. The open-vocabulary relationship classifier leverages the rich semantic knowledge of CLIP to predict relationships between the generated object trajectories.
By jointly training the trajectory detector and the relationship classifier, our framework does not suffer from the domain gap problem faced by existing methods that rely on pre-trained trajectory detectors. Moreover, by distilling the visual encoder of CLIP, our framework enhances the generalization to novel object categories thanks to CLIP's powerful representation capabilities. 

To exploit the relationship context during trajectory detection, we propose to embed a relationship query into the query-based Transformer decoder, and design an auxiliary relationship loss accordingly to explicitly perceive the relationships between objects when decoding. 
By incorporating relationship context into trajectory detection, our framework enables mutual interactions between trajectory detection and relationship classification.
This mutual interaction fosters a close coupling that facilitates the joint optimization of both processes within the end-to-end framework, ensuring that object trajectories and relationships are simultaneously refined and accurately detected.

To effectively leverage the knowledge of CLIP into the video domain during relationship classification, we propose a multi-modal prompting method that prompts CLIP on both visual and language sides.
Specifically, we design spatio-temporal visual prompting to imbue CLIP with the capabilities of spatial and temporal modeling, effectively enhancing the image encoder of CLIP. Moreover, we design vision-guided language
prompting to exploit CLIP’s comprehensive semantic knowledge for discovering novel relationships in videos.

Extensive experiments on two public datasets, VidVRD~\cite{shang2017video} and VidOR~\cite{shang2019relation}, show that our end-to-end framework outperforms existing state-of-the-art methods, achieving 2.89\% mAP gains on novel relationship categories on the VidVRD dataset. 
To further demonstrate the generalization ability of our method, our framework is also applied to a more difficult evaluation setting where the base categories of VidOR are used for training and unseen categories from VidVRD are used for testing. 
Under this setting, our framework achieves 9.77\% mAP improvement on trajectory detection and 5.45\% mAP improvement on relationship classification.

In summary, our main contributions are as follows:
\begin{enumerate}
    \item We propose an end-to-end Open-VidVRD framework, which unifies trajectory detection and relationship classification, thus eliminating the need for pre-trained trajectory detectors and improving the generation to unseen categories. 
    \item We propose a relationship-aware open-vocabulary trajectory detector, which distills significant knowledge from CLIP and meanwhile perceives the relationship context via a dedicated relationship query and an auxiliary relationship loss.
    \item We also propose an open-vocabulary relationship classifier with a multi-modal prompting method that prompts CLIP on both the visual and language sides to enhance the generalization to novel relationship categories.
\end{enumerate}

A preliminary version of this paper, named OV-MMP~\cite{yang2024multi}, published in AAAI 2024. The differences between this paper and the previous version are summarized as follows: (1) This paper integrates the open-vocabulary relationship classifier with the multi-modal prompting method into a novel end-to-end framework, eliminating the reliance on pre-trained trajectory detector used in OV-MMP, enabling the joint optimization of trajectory detection and relationship classification. (2) This paper proposes a relationship-aware open-vocabulary trajectory detector that distills significant knowledge from CLIP visual encoder into the query-based Transformer decoder while explicitly perceiving the relationship context by designing a relationship query and an auxiliary relationship loss. (3) This paper further validates the generalization capability of our end-to-end framework by designing extensive experiments, including a new setting in which we train the model on the base categories of the VidOR dataset and test it on categories that are unseen during training from the VidVRD dataset.

\section{Related Work}
\label{sec:related_work}
\subsection{Video Visual Relationship Detection}
Video Visual Relationship Detection (VidVRD) focuses on detecting interactions between objects over time, necessitating a comprehensive understanding of both spatial distribution and temporal dynamics of objects within videos~\cite{shang2017video}. 
Numerous studies have explored various VidVRD methods, which can be broadly categorized into spatio-temporal modeling, relationship refinement, video relationship debiasing, and end-to-end video relationship detection.

Spatio-temporal modeling methods design various architectures to learn dynamic interactions between objects across both spatial and temporal dimensions. 
Qian et al.~\cite{qian2019video}, Tsai et al.~\cite{tsai2019video}, and Liu et al.~\cite{liu2020beyond} represent videos as fully connected spatio-temporal graphs and adopt graph convolution networks to reason about the relationships between objects.
Cong et al.~\cite{cong2021spatial} use a spatial Transformer encoder to extract spatial context and intra-frame relationships, and a temporal decoder to understand inter-frame dynamic relationships. 

Relationship refinement methods aim to learn fine-grained relationship representation between objects. 
Shang et al.~\cite{shang2021video} propose an iterative inference module that iteratively refines one component of a relationship triplet by using the prediction results of the other two components.
Chen et al.~\cite{chen2021social} decouple complex relationships across multiple video frames into fine-grained relationships on single frames to capture frame-wise subtle interactions between objects.

Video relationship debiasing methods aim to address the long-tail distribution problem in video relationship datasets.
Xu et al.~\cite{xu2022meta} apply meta-learning to train an unbiased VidVRD model. 
They divide the training set into a support set and multiple query sets with different data distributions, where the support set is used to train the model, the query sets are used to optimize the model. 
Dong et al.~\cite{dong2022stacked} divide long-tail datasets into balanced sub-datasets, and individually train a relationship classifier for each subset. Then, they jointly optimize the classifiers on the full training set and distill the unbiased knowledge in each classifier into a comprehensive classifier.
Lin et al.~\cite{lin2024td2} design an asymmetrical re-weighting loss function that adjusts the weights for each relationship category by using the effective number of samples proposed in~\cite{cui2019class}.

End-to-end video relationship detection methods~\cite{zheng2022vrdformer,zhang2023end} have been proposed in recent years. They jointly optimize both the trajectory detector and relationship classifier to improve the consistency of the object and relationship context.

All the above-mentioned methods are designed for closed settings where the training and test data share the same object and relationship categories, thus limiting their ability to generalize to unseen object and relationship categories. Consequently, they struggle to effectively adapt to the diverse and dynamic scenarios encountered in real-world videos.

\subsection{Open-vocabulary Visual Relationship Detection}
{With the advancement of vision-language pre-training techniques, open-vocabulary visual tasks, such as object detection~\cite{ren2024dino}, spatio-temporal action localization~\cite{wu2024open}, and video-text matching~\cite{wu2024building}, have gained widespread attention for their ability to generalize beyond pre-defined categories.}

The task of open-vocabulary visual relationship detection~\cite{he2022towards, gao2023compositional} has been proposed recently, which focuses on detecting visual relationship instances involving objects and relationships that are unseen in the training data. 
The first study~\cite{he2022towards} on this task conducts contrastive learning on massive amounts of data to align the visual and textual representations of both objects and relationships, and identifies novel categories through similarity matching between visual content and textual descriptions.
{Further advancing this approach, Yuan et al.~\cite{yuan2023rlipv2} propose to enhance relational language-image pre-training by accelerating convergence through early cross-modal fusion and improving scalability with pseudo-labeled relational annotations.}

Due to the high computational demands of contrastive learning with large-scale data, many researchers resort to using rich semantic knowledge in existing pre-trained vision-language models~\cite{li2022blip,radford2021learning,li2022align} to recognize novel categories. 
Li et al.~\cite{li2024pixels} use BLIP~\cite{li2022blip} to generate relationship triplets by feeding images and text prompts, and then replace synonyms in the generated triplets with the target categories through text similarity matching. 
Li et al.~\cite{li2024zero} enhance CLIP~\cite{radford2021learning} to discover novel relationships by generating fine-grained visual and textual content. Specifically, they use object detection results to decompose the visual content into subject-related, object-related, and spatial-related fine-grained components, and adopt large language models (LLMs) to generate class-specific descriptive prompts for each component. 
Yu et al.~\cite{yu2023visually} also use CLIP for open-vocabulary relationship classification, and propose a prompting method that concatenates learnable vectors to both textual and visual input to learn task-related knowledge. Moreover, they take the visual features and learnable text prompts into BERT~\cite{devlin2018bert} to generate comprehensive and fine-grained object relationships for expanding the training data.  
Zhao et al.~\cite{zhao2023unified} unify inconsistent label spaces across multiple datasets by leveraging the aligned vision-text semantic space in CLIP. 
Zhu et al.~\cite{zhu2024towards} employ a mask-based approach to unify multiple relationship understanding tasks, using CLIP text prompts to guide visual relationship segmentation and a query-based Transformer to generate relational triplets.

The above-mentioned methods are typically designed for images and can not be directly applied to video domains. 
In recent years, Gao et al.~\cite{gao2023compositional} pioneer open-vocabulary video visual relationship detection (Open-VidVRD) by using the pre-trained video-text model ALPro~\cite{li2022align} for similarity matching between visual and linguistic modality features. However, this method relies heavily on a trajectory detector pre-trained on closed datasets, thus limiting the ability to generalize to unseen object categories.

In this paper, we propose a novel end-to-end Open-VidVRD framework that jointly models object trajectory detection and relationship classification, eliminating the reliance on pre-trained trajectory detectors. Moreover, by leveraging the knowledge in CLIP through a novel multi-modal prompting method, our framework adapts well to diverse real-world scenarios.

\subsection{Prompting CLIP}

Vision-language pre-trained models~\cite{radford2021learning, alayrac2022flamingo, li2020hero, luo2020univl} have demonstrated significant progress in many downstream vision-language tasks. 
As one of the most successful vision-language pre-trained models, CLIP~\cite{radford2021learning}, is extensively pre-trained using 400 million image-text pairs from the Internet, resulting in a vision-language embedding space with comprehensive semantic knowledge.

Various text prompting methods have emerged to effectively transfer knowledge from CLIP to downstream tasks. 
Zhou et al.~\cite{zhou2022learning} convert handcraft text prompts into learnable vectors to learn task-related knowledge. 
Zhou et al.~\cite{zhou2022conditional} further propose conditional text prompts, which integrate learnable vectors with visual features, to learn the image-specific knowledge. 
Sun et al.~\cite{sun2022dualcoop} adapt CLIP to a multi-label image recognition task by learning pairs of positive and negative text prompts to ensure independent binary classification for each category.

Meanwhile, many visual prompting methods for CLIP have been widely explored. 
Jia et al.~\cite{jia2022visual} integrate the input images with learnable vectors to learn task-related visual cues. 
Wang et al.~\cite{wang2022learning} and Xu et al.~\cite{xu2023progressive} incorporate learnable tokens into the visual encoder to refine the visual features, making them more suitable for downstream tasks. 

To fully exploit the multi-modal co-optimization potential of CLIP, multi-modal prompting methods~\cite{khattak2023maple,li2023efficient,xin2024mmap} have been proposed. 
These methods introduce learnable vectors to both visual and textual modalities, and couple them to facilitate joint optimization.
All these methods primarily focus on image domain tasks. In contrast, our multi-modal prompting method is specifically designed for the more challenging Open-VidVRD task.

\begin{figure*}[h]
\centering
\includegraphics[width=0.88\textwidth]{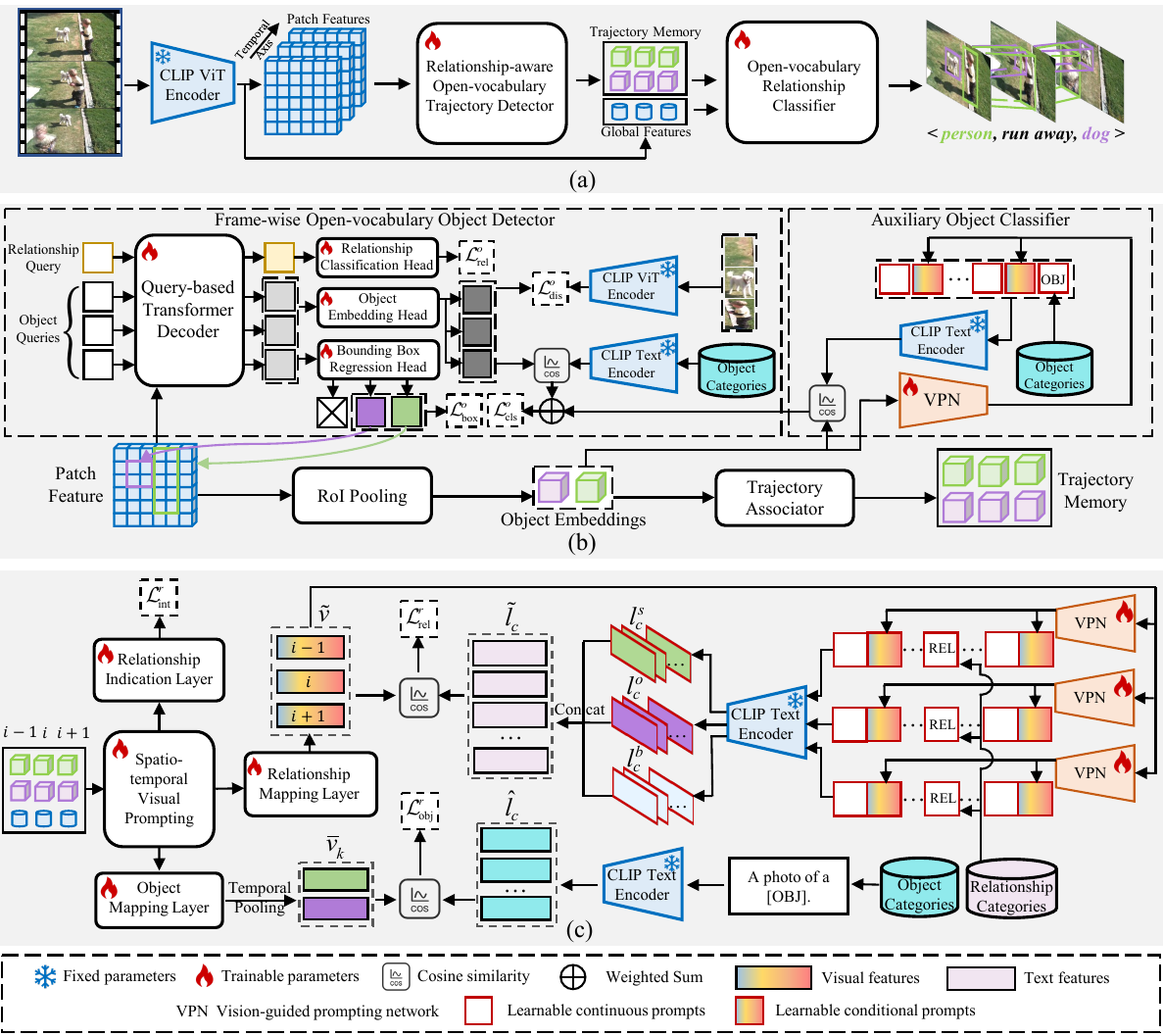}
\caption{{
(a) The proposed end-to-end framework, where the object trajectories and their categories are predicted by the relationship-aware open-vocabulary trajectory detector, and the relationship categories are predicted by the open-vocabulary relationship classifier. (b) The relationship-aware open-vocabulary trajectory detector. (c) The open-vocabulary relationship classifier. }} 
\label{fig:overview}
\end{figure*}

\section{Our framework}
\label{sec:method}
\subsection{Overview}
\label{sec:overview}
Video Visual Relationship Detection (VidVRD) aims to detect instances of visual relationships within a video $V=\{f_t\}_{t=1}^{N_v}$, where $f_t$ represents the frame at time  $t$, and $N_v$ is the number of frames in $V$. 
Each visual relationship instance is represented by a tuple $(c^s,c^r,c^o,{T}^s,{T}^o)$, where $c^s$, $c^r$, and $c^o$ denote the categories of the subject, relationship, and object, respectively. 
${T}^k$, with $k \in \{s, o\}$, represents the trajectory of subjects or objects, comprising a sequence of bounding boxes $(b_{t_{s}}^k,\dots, b_{t_{j}}^k,\dots, b_{t_{e}}^k)$, where $b_{t_j}^k$ denotes the corresponding bounding box at time $t_j$, with $t_{s}$ and $t_{e}$ being the start time and end time of the trajectory, respectively.
In Open-VidVRD, the categories of objects and relationships are divided into base and novel splits: base objects ($\mathcal{C}_b^O$), novel objects ($\mathcal{C}_n^O$), base relationships ($\mathcal{C}_b^R$), and novel relationships ($\mathcal{C}_n^R$). 
Only base categories are used during the training phase, and all categories are used during the test phase.

We propose an end-to-end Open-VidVRD framework that directly detects relationships between objects from input raw videos. 
It comprises two main components: a relationship-aware open-vocabulary trajectory detector (Sec.~\ref{sec:object}) and an open-vocabulary relationship classifier (Sec.~\ref{sec:relationship}). 
An overview of our framework is illustrated in Figure~\ref{fig:overview} (a).

\subsection{Relationship-aware Open-vocabulary Trajectory Detection}
\label{sec:object}
For each input video $V$, we first use a ViT-based visual encoder~\cite{dosovitskiy2020image} of CLIP to extract visual features for each frame, represented by
\begin{equation}
\label{eq:global}
    (F_t^g,F_t^p)=\mathcal{V}(V),
\end{equation}
where $\mathcal{V}(\cdot)$ denotes the visual encoder of CLIP, $F_t^g$ and $F_t^p$ denote the global feature and the patch feature of the $t$-th frame, respectively. 
We then feed the patch features into a relationship-aware open-vocabulary trajectory detector to obtain object trajectories, represented by
\begin{equation}
(T_i,c_i,E_i)=\Phi(F_1^p,F_2^p,\dots,F_{N_v}^p),
\end{equation}
where $\Phi(\cdot)$ denotes the  trajectory detector. 
$T_i$ is the $i$-th trajectory, where $i \in \{1,...,N_t\}$ and $N_t$ is the trajectory number in the video. 
$c_i$ denotes the object category of the $i$-th trajectory. $E_i$ represents the visual feature of the $i$-th trajectory. 

The trajectory detection process begins with a query-based Transformer decoder that distills the visual encoder of CLIP to perform frame-wise open-vocabulary object detection (Sec.~\ref{sec:detr}). 
Then the frame-wise object detection results are enhanced by an auxiliary object classifier (Sec.~\ref{sec:object_classification}) that leverages CLIP to discover novel object categories.
Finally, a trajectory associator (Sec.~\ref{sec:association}) connects the frame-wise detection results to generate coherent object trajectories throughout the video. 
A trajectory memory is built to store the trajectory detection results, which are then used for subsequent open-vocabulary relationship classification.

We further propose a relationship query as input for the query-based Transformer decoder and design a corresponding auxiliary relationship loss (Sec.~\ref{sec:loss1}) to make the decoder explicitly perceive the relationship context. 
Figure~\ref{fig:overview} (b) illustrates the details of the proposed relationship-aware open-vocabulary trajectory detector.  

\subsubsection{Frame-wise Open-vocabulary Object Detection}
\label{sec:detr}
We feed the patch feature together with object queries and a relationship query into the query-based Transformer decoder~\cite{carion2020end} to obtain the query results. The object query results are then processed through prediction heads to generate frame-wise object detection results, and the relationship query result is used to calculate the auxiliary relationship loss.

\textbf{Object Query.}
We define a set of \( N_q \) learnable object queries, denoted by \({Q}=\{{q}^1, {q}^2, \ldots, {q}^{N_q}\}\), to process image context and output predictions in parallel. The object queries are shared across all video frames.

\textbf{Relationship Query.}
We propose a relationship query, denoted as \( R \), which interacts with the patch feature ${F}^p$ to perceive the relationship context. The relationship query is shared across all video frames.

\textbf{Query-based Transformer Decoder.}
The query-based Transformer decoder has \(N_l\) layers, and each layer is composed of alternating self-attention and cross-attention modules. In the \( l \)-th layer, the object queries \({Q}^l\) and relationship query \({R}^l\)  first interact with each other through the self-attention module. The resulting outputs, \(\tilde{Q}^l\) and \(\tilde{R}^l\), are used to extract features from the patch feature \({F}^{p}\) through the cross-attention module. This process is formulated as
\begin{equation}
\begin{gathered}
(\tilde{Q}^l, \tilde{R}^l ) = \text{SelfAttn}_l\left([{Q}^l; {R}^l]\right), \\
(\hat{Q}^l, \hat{R}^l ) = \text{CrossAttn}_l\left([\tilde{Q}^l;\tilde{R}^l], {F}^{p}\right),
\label{eq:decoder}
\end{gathered}
\end{equation}
where \(\hat{Q}^l\) and \(\hat{R}^l\) represent the output of the \( l \)-th decoder layer, and serve as the input \({Q}^{l+1}\) and \({R}^{l+1}\) for the $(l+1)$-th layer. 
The final output of the query-based Transformer decoder is represented by \(\hat{Q}^{N_l} = \{\hat{q}^1, \hat{q}^2, \ldots, \hat{q}^{N_q}\}\) and \(\hat{R}^{N_l} \).

\textbf{Prediction Heads.}
For each object query result \(\hat{q} \in \hat{Q}^{N_l}\), its corresponding object bounding box is predicted as
\begin{equation}
    b = \mathcal{M}_{box}({\hat{q}}),
\end{equation}
where $b$ is the predicted bounding box, \(\mathcal{M}_{box}(\cdot)\) is the bounding box regression head, consisting of two linear layers. The classification score of object category \( c \in C \) is represented as
\begin{equation}
\label{eq:mc}
    p(c)=\frac{\text{exp}(\text{cos}(\mathcal{M}_{emb}({\hat{q}}), {e}^{c}_\text{txt})/\tau)}{ {\textstyle \sum_{c'\in C} \text{exp}(\text{cos}(\mathcal{M}_{emb}({\hat{q}}),  {e}^{c'}_\text{txt})/\tau)}},
\end{equation}
where \( C = \mathcal{C}_b^{O} \) during the training phase and \( C = \mathcal{C}_b^{O} \cup \mathcal{C}_n^{O} \) during the test phase,  \(\tau\) is a temperature parameter, \(\mathcal{M}_{emb}(\cdot)\) is the object embedding head consisting of two linear layers,  \(\text{cos}(\cdot,\cdot)\) is the cosine similarity matching function, and \({e}^{c}_\text{txt}\) denotes the text feature of object category \( c \) extracted by CLIP. We retain only the bounding boxes where the maximum value of \(p\) exceeds a threshold \(\epsilon\) (an ablation is presented in Table~\ref{tab:epo}), and discard the other bounding boxes. For the retained bounding boxes, we create a mask \({M}\) and apply Region of Interest (RoI) Pooling on the original patch feature \({F}^{p}\) to extract a fixed-size feature vector as the object embedding \(\mathcal{E}\),  formulated as
\begin{equation}
\label{epxl}
    \mathcal{E}=\mathcal{P}({F}^{p},{M}),
\end{equation}
where \(\mathcal{P}(\cdot)\) denotes the RoI Pooling function.

For the relationship query result \(\hat{R}^{N_l}\), we predict the score of relationship category \( r \in \mathcal{C}_b^R \)  by
\begin{equation}
p(r)=\mathcal{M}_{rel}(\hat{R}^{N_l}),
\end{equation}
where \(\mathcal{M}_{rel}(\cdot)\) is the relationship classification head, consisting of two linear layers.

\subsubsection{Auxiliary Object Classification}
\label{sec:object_classification}
To further improve the classification performance of novel object categories, we design an auxiliary object classifier that uses CLIP to classify the object embeddings obtained in Eq.~\ref{epxl} by calculating their similarity with text features.
To fully leverage the rich semantic knowledge of CLIP, we propose a vision-guided prompting method.
Specifically, we feed the object embeddings into a vision-guided prompting network (VPN) to generate learnable conditional language prompts. These generated prompts are then combined with learnable continuous language prompts as input for the text encoder of CLIP.

\textbf{Learnable Continuous Language Prompts.}
For each object category [OBJ],  $N_{\varsigma}$-token language prompts are initialized by $\varsigma = [\boldsymbol{\varsigma}_1, \boldsymbol{\varsigma}_2,\cdots, \boldsymbol{\varsigma}_{N_{\varsigma}}]$, where $\text{[OBJ]} \in \mathcal{C}_b^O$ when training and $\text{[OBJ]} \in \mathcal{C}_n^O \cup \mathcal{C}_b^O$ when testing,

\textbf{Learnable Conditional Language Prompts.}
For each object category [OBJ], $N_{\zeta}$-token learnable conditional language prompts are learned by taking into account the corresponding visual feature,  represented as
\begin{equation}
    \boldsymbol{\zeta} = [\boldsymbol{\zeta}_1, \boldsymbol{\zeta}_2,\cdots, \boldsymbol{\zeta}_{N_{\zeta}}] = \varphi(\mathcal{E}), 
\end{equation}
where $\varphi(\cdot)$ denotes the vision-guided prompting network, consisting of two linear layers. $\mathcal{E}$ is the object embedding.

\textbf{Learnable Vision-guided Language Prompts.}
We concatenate the tokens of learnable continuous language prompts and tokens of learnable conditional language prompts interlaced, and then insert the [OBJ] token into the end of the token sequence, to obtain the final language prompts $\boldsymbol{\jmath}_{\text{OBJ}} = \left [\boldsymbol{\varsigma}_1, \boldsymbol{\zeta}_1,\boldsymbol{\varsigma}_2, \boldsymbol{\zeta}_2,\cdots,\boldsymbol{\varsigma}_{N_{\varsigma}}, \boldsymbol{\zeta}_{N_{\zeta}}, \text{OBJ}  \right]$. 
The text feature of the object category  $c$ is denoted as
\begin{equation}
    \boldsymbol{j}_{c} = \mathcal{T}(\boldsymbol{\jmath}_c), 
\end{equation}
where $\mathcal{T}(\cdot)$ is the text encoder of CLIP.

The auxiliary classification score of object category \( c \in C \) is represented as
\begin{equation}
\label{eq:mc2}
    \tilde{p}(c)=\frac{\text{exp}(\text{cos}(\mathcal{E}, \boldsymbol{j}_{c})/\tau)}{ {\textstyle \sum_{c'\in C} \text{exp}(\text{cos}(\mathcal{E},  \boldsymbol{j}_{c'})/\tau)}}.
\end{equation}
The final frame-wise object classification score is represented as
\begin{equation}
\label{eq:weighting1}
\hat{p}(c)=
\left\{
\begin{aligned}
    &(1-\alpha) p(c)+\alpha\tilde{p}(c) \:\text{if} \: c \in \mathcal{C}_b^O,\\
    &(1-\beta )p(c)+\beta \tilde{p}(c) \:\text{if} \: c \in \mathcal{C}_n^O,
\end{aligned}
\right.
\end{equation}
where $\alpha,\beta \in [0,1]$ are weighting factors for the base and novel object categories, respectively.

\subsubsection{Trajectory Association}
\label{sec:association}
We employ a feature-based association algorithm~\cite{wojke2017simple} that links frame-wise detection results that are spatially close and visually similar to generate object trajectories $(T_1,T_2,...,T_{N_t})$, where $N_t$ is the number of trajectories in the video. 

For the $i$-th trajectory, its object classification score \( P_i \) is calculated by averaging the final frame-wise object classification scores,  formulated as
\begin{equation}
\label{equ:score}
\hat{P}_{i} = \frac{1}{t_{\text{e}} - t_{\text{s}} + 1} \sum_{t=t_{\text{s}}}^{t_{\text{e}}} \hat{p}_t^i,
\end{equation}
where $\hat{p}_t^i$ represents the final classification score of the $i$-th trajectory at the $t$-th frame, $t_s$ and $t_e$ are the start and end time of the trajectory, respectively. 
The predicted object category of the $i$-th trajectory is given by 
\begin{equation}
    c_i = \arg \max_{c} \hat{P}_i(c).
\end{equation}

The visual feature of the $i$-th trajectory is represented as
\begin{equation}
\label{eq:feature}
    {E}_i = \{\mathcal{E}_t^i\}_{t=t_{\text{s}}}^{t_{\text{e}}},
\end{equation}
where \(\mathcal{E}_t^i\) is the object embedding of the $i$-th trajectory at the $t$-th frame. $(T_i,c_i,E_i)$ are stored in the trajectory memory for subsequent relationship classification. 
\subsubsection{Training Loss}
\label{sec:loss1}
We use a focal loss~\cite{lin2017focal} for object classification, and an L1 loss and a GIoU loss~\cite{rezatofighi2019generalized} for box regression. 

\textbf{Distillation Loss.}
We design a distillation loss to distill the knowledge from CLIP's visual encoder for frame-wise open-vocabulary object detection, formulated as
\begin{equation}
\mathcal{L}^o_{dis} = \frac{1}{N_b}\cdot\sum\nolimits_{n=1}^{N_b}\|\textbf{e}^n-\textbf{z}^n\|_1,
\end{equation}
where $\mathcal{L}^o_{dis}$ represents the distillation loss, $N_b$ denotes the total number of the retained bounding boxes in the video, $\textbf{e}^n$ is the visual feature of the $n$-th bounding box extracted by CLIP, and $\textbf{z}^n$ is the corresponding object query result encoded by the object embedding head, \textit{i.e.}, $\mathcal{M}_{emb}({\hat{q}})$ in Eq.~\ref{eq:mc}.

\textbf{Auxiliary Relationship Loss.}
We propose an auxiliary relationship loss, calculated using Binary Cross-Entropy (BCE), to enable the decoder to explicitly perceive the relationships between objects. 
The object relationships are categorized  into dynamic and static types. Dynamic relationships change significantly over time and require multiple video frames to be assessed together to make a judgment. For example, a dynamic relationship such as ``run past" involves motion that unfolds over multiple frames, which means that the relationship between objects can only be understood by analyzing how the scene evolves over time. 
In contrast, static relationships remain relatively constant and can be determined from a single video frame. For example, a static relationship such as ``lie behind" describes a stable spatial configuration between objects that does not require temporal tracking. Once the relative position is established in a single frame, the relationship remains clear.

To emphasize the importance of static categories for understanding frame-wise relationships, we adjust the BCE loss with a predefined weight \(\lambda_s\), ensuring that static relationships receive appropriate emphasis in the learning process.
The auxiliary relationship loss \(\mathcal{L}^o_{rel}\) is formulated as

\begin{equation}
\label{eq:weighting3}
\begin{gathered}
\mathcal{L}^o_{rel} = \mathcal{L}_{d} + \lambda_s\mathcal{L}_{s},\\
\mathcal{L}_{d} = \frac{1}{N_v}\cdot \sum\nolimits_{t=1}^{N_v}\text{BCE}(r_t^{d},\hat{r}_t^d),\\
\mathcal{L}_{s} = \frac{1}{N_v}\cdot \sum\nolimits_{t=1}^{N_v}\text{BCE}(r_t^{s},\hat{r}_t^s),
\end{gathered}
\end{equation}
where $r_t^d$ and $r_t^s$ represent the predicted scores of dynamic and static relationship categories in the $t$-th video frame, respectively, and  $\hat{r}_t^d$ and $\hat{r}_t^s$ represent the ground-truth labels of dynamic and static relationship categories, respectively. $N_v$ is the number of frames. 

The overall training loss of our relationship-aware open-vocabulary trajectory detector is given by
\begin{equation}
\label{eq:weighting2}
\mathcal{L}^o =  \lambda_1\mathcal{L}^o_{foc} + \lambda_2\mathcal{L}^o_{l1} + 
\lambda_3\mathcal{L}^o_{iou} + \lambda_4\mathcal{L}^o_{dis} + \lambda_5\mathcal{L}^o_{rel},
\end{equation}
where $\mathcal{L}^o_{foc}$, $\mathcal{L}^o_{l1}$, $\mathcal{L}^o_{iou}$, $\mathcal{L}^o_{dis}$, and $\mathcal{L}^o_{rel}$ represent the focal loss, L1 loss, GIoU loss, distillation loss and auxiliary relationship loss, respectively. 
Note that all the aforementioned losses are calculated using the frame-level object detection results.

\subsection{Open-vocabulary Relationship Classification}
\label{sec:relationship}
We pair the detected object trajectories with temporal overlap in the trajectory memory and denote each trajectory pair as \((T^s, c^s, E^s, T^o, c^o, E^o)\), where \(T^s\), \(c^s\), and \(E^s\) represent the trajectory, object category, and visual feature of the subject in the trajectory pair, respectively, and \(T^o\), \(c^o\), and \(E^o\) represent the trajectory, object category, and visual feature of the object in the trajectory pair.
We extract the visual features of the background of trajectory pair by aggregating the global feature of each frame, denoted as ${E}^h=\{F^g_{t}\}_{t=t_{\text{s}}}^{t_{\text{e}}}$, where \( F^g_{t} \) is the global visual feature (extracted in Eq.~\ref{eq:global}) of the \( t \)-th video frame, $t_s$ and $t_e$ are the start and end frames of the trajectory pair.
Then we feed the visual features of the subject, object, and background into an open-vocabulary relationship classifier which uses CLIP to generate the relationship classification results by calculating the similarity between the visual features and text features, represented by
\begin{equation}
    c^r=\Psi(E^{s},E^{o},E^h),
\end{equation}
where  \(c^r\) represents the predicted relationship category label, \(\Psi(\cdot)\) represents the open-vocabulary relationship classifier.

To adapt CLIP well to relationship classification, we propose a multi-modal prompting method that applies prompt learning to both visual and textual branches of CLIP. 
Specifically, we propose a spatio-temporal visual prompting method (Sec.~\ref{sec:visual}) to capture dynamic contexts, and a vision-guided language prompting method (Sec.~\ref{sec:language}) to exploit CLIP’s comprehensive semantic
knowledge for discovering unseen relationship categories.
Figure~\ref{fig:overview} (c) illustrates the details of the proposed open-vocabulary relationship classifier.  
\subsubsection{Spatio-temporal Visual Prompting}
\label{sec:visual}

We use standard Transformer blocks to model the sptaio-temporal relationships between objects.
To reduce the computational complexity, we decouple the spatio-temporal modeling into separate and successive modules, namely spatial modeling and temporal modeling.

\textbf{Spatial Modeling.}
Spatial relationships between objects are typically defined by their positional orientations, such as being in front of or above each other. Therefore, spatial modeling requires combining three key elements: features of the subject region, features of the object region, and features representing the background (\textit{i.e.} the whole image). This process involves modeling interactions between objects and their background to capture spatial context, thus enhancing object features.

Given the features of the trajectory pair, denoted by \({E}^k, k \in \{s, o, h\}\), we add two types of learnable embeddings: positional embedding \(\varrho^k\) related to the normalized bounding box, and role embedding \(\rho^k\). These two types of embeddings are learned and shared across all video frames. 
The visual features are updated as follows:
\begin{equation}
	(\boldsymbol{\dot{v}}^s, \boldsymbol{\dot{v}}^o, \boldsymbol{\dot{v}}^h) = {\rm STrans}(\boldsymbol{I}^s, \boldsymbol{I}^o, \boldsymbol{I}^h),
\end{equation}
where \(\boldsymbol{I}^k = {E}^k + \varrho^k + \rho^k, k \in \{s, o, h\}\), and STrans($\cdot$) denotes the spatial Transformer blocks.

\textbf{Temporal Modeling.}
Temporal relationships of objects are time-dependent, such as moving toward or away, so the inputs for temporal modeling include visual features and temporal embeddings. For simplicity, the same temporal modeling is applied to different roles, \textit{i.e.}, subject, object, and their background in this paper. Through the exploration of dynamic state transformations, the visual features are systematically updated.

Given the spatially encoded visual features \(\boldsymbol{\dot{v}}=\{\boldsymbol{\dot{v}}_t^s,\boldsymbol{\dot{v}}_t^o,\boldsymbol{\dot{v}}_t^h\}_{t=t_{\text{s}}}^{t_{\text{e}}}\), for each role, we collect the corresponding features across all frames, denoted as \(\boldsymbol{\dot{v}}^k=\{\boldsymbol{\dot{v}}_t^k\}_{t=t_{\text{s}}}^{t_{\text{e}}}\), where \( k \in \{s, o, h\} \). We then add temporal embedding \(\theta_t\), which is related to frame \( t \) and shared across all roles. For each role, the visual features are updated by
\begin{equation}
	\boldsymbol{\ddot{v}}^k = \{\boldsymbol{\ddot{v}}_t^k\}_{t=t_{\text{s}}}^{t_{\text{e}}} = {\rm TTrans}(\boldsymbol{\dot{I}}_{t_{\text{s}}}^k, \boldsymbol{\dot{I}}_{t_{\text{s}}+1}^k, \cdots, \boldsymbol{\dot{I}}_{t_{\text{e}}}^k),
\end{equation}
where \(\boldsymbol{\dot{I}}_t^k = \boldsymbol{\dot{v}}_t^k + \theta_t\), and TTrans($\cdot$) denotes the temporal Transformer blocks.

\subsubsection{Vision-guided Language Prompting}
\label{sec:language}
Similar to Sec.~\ref{sec:object_classification}, we construct vision-guided language prompts as
\begin{equation}
\boldsymbol{\ell}_{\text{REL}}^k = \left [\boldsymbol{\varsigma}_1^k, \boldsymbol{\zeta}_1^k,\boldsymbol{\varsigma}_2^k, \boldsymbol{\zeta}_2^k,\cdots, \text{REL}, \cdots, \boldsymbol{\varsigma}_{N_{\varsigma}}^k, \boldsymbol{\zeta}_{N_{\zeta}}^k \right],  
\end{equation}
where \( k \in \{s, o, h\} \) and \(\text{[REL]} \in \mathcal{C}_b^R\) during training and \(\text{[REL]} \in \mathcal{C}_n^R \cup \mathcal{C}_b^R\) during testing.
For each visual region, the final text features of relationship category \( r \)  are given by
\begin{equation}
    \boldsymbol{l}_{r}^k = \mathcal{T}(\boldsymbol{\ell}_r^k),
\end{equation}
where \(\mathcal{T}(\cdot)\) is the text encoder of CLIP.

\subsubsection{Training Loss}
\label{sec:loss2}
{The training loss of the open-vocabulary relationship classifier consists of three parts: a relationship classification loss $\mathcal{L}^r_{rel}$, an object classification loss $\mathcal{L}^r_{obj}$, and an interaction loss $\mathcal{L}^r_{int}$, as shown in Figure \ref{fig:overview} (c). }
The overall training loss is given by
\begin{equation}
\label{eq:weighting4}
\mathcal{L}^r = \mathcal{L}^r_{rel} + \gamma\mathcal{L}^r_{obj} + \delta\mathcal{L}^r_{int}.
\end{equation}

{\textbf{Relationship Classification Loss.}}
Given the visual features $\boldsymbol{\ddot{v}}^{k}$ and the text features $\boldsymbol{l}_{r}^k$, the prediction score of the relationship category $r$ is calculated by
\begin{equation}    
\hat{y}_{r}^{rel} = \sigma(\cos(\boldsymbol{\Tilde{v}},  \boldsymbol{\Tilde{l}_r})),
\end{equation}
where $\boldsymbol{\Tilde{v}} = \psi([\boldsymbol{\ddot{v}}^{s};\boldsymbol{\ddot{v}}^{o};\boldsymbol{\ddot{v}}^{h}])$, $\psi(\cdot)$ denotes the relationship mapping layer in Figure~\ref{fig:overview} (c), $\boldsymbol{\Tilde{l}_r} = [\boldsymbol{l}_r^{s};\boldsymbol{l}_r^{o};\boldsymbol{l}_r^{h}]$, 
$\sigma(\cdot)$ is the sigmoid function, $\cos(\cdot,\cdot)$ is the cosine similarity.
{The relationship classification loss is formulated by using the BCE loss:}
\begin{equation}
\mathcal{L}^r_{rel} = \frac{1}{|\mathcal{C}_b^R|}\cdot \sum\nolimits_{r \in \mathcal{C}_b^R}{\rm BCE}(\hat{y}_{r}^{rel}, y_{r}^{rel}),
\end{equation}
where $y_r^{rel} = 1$ when  $r$ equals to the ground-truth relationship category, otherwise  $y_r^{rel} = 0$.

{\textbf{Object Classification Loss.} To avoid the visual feature drift caused by spatio-temporal visual prompting, we introduce an object classification loss to enforce the visual features after spatial modeling to have the same object distinguishing capability as the original CLIP. }
Specifically, after the spatial modeling, we collect the subject and object features from all frames and average them as $\boldsymbol{\bar{v}}^k={\rm avg}(\{\phi(\boldsymbol{\dot{v}}_t^k)\}_{t=0}^{T})$, $k \in \{s,o\}$ and $\phi(\cdot)$ denotes the object mapping layer, as shown in Figure~\ref{fig:overview} (c). Meanwhile, we extract the text features for all subject or object categories by feeding the handcrafted prompts (\textit{i.e.}, ``a photo of [OBJ]”) into the text encoder of CLIP, where [OBJ] can be replaced with the names of subjects or objects. 
The similarity between the visual features and the text features of object category $c$ is calculated by $\hat{y}_c^k = \cos(\boldsymbol{\bar{v}}^k,  \boldsymbol{\hat{l}}_c), k\in \{s,o\}$.
{Finally, the object classification loss is computed over all object categories using the cross-entropy loss (CE):}
\begin{equation}
    \mathcal{L}^r_{obj} = {\rm CE}(\boldsymbol{\hat{y}}^{s}, {y}^{s}) + {\rm CE}(\boldsymbol{\hat{y}}^{o}, {y}^{o}),
\end{equation}
where $\boldsymbol{\hat{y}}^{s}$ is the predicted subject similarity between visual features and text features of base object categories ($\mathcal{C}_b^O$), and $\boldsymbol{\hat{y}}^{o}$ is the corresponding predicted object similarity. ${y}^{s}$ and $ {y}^{o}$ denote the ground-truth category labels of the subject and object, respectively.

\textbf{Interaction Loss.}
There may be no annotated relationships between some subjects and objects, that is, there is no interaction. For each pair of subject and object, if there are any relationship categories between them in video frame $t$, we set the ground-truth interaction by $y_t^{int}=1$, otherwise $y_t^{int}=0$.  To learn this weak interaction,  we concatenate all the features in frame $t$ and predict the interaction probability by $\hat{y}_t^{int} = \psi([\boldsymbol{\ddot{v}}_t^{s};\boldsymbol{\ddot{v}}_t^{o};\boldsymbol{\ddot{v}}_t^{h}])$, where $\psi(\cdot)$ denotes the relationship indication layer in Figure~\ref{fig:overview} (c). 
The interaction loss is then computed using the binary cross-entropy loss (BCE):
\begin{equation}
\mathcal{L}_{int} = \frac{1}{t_e-t_s} \cdot \sum\nolimits_{t=t_s}^{t_e}{\text{BCE}(\hat{y}^{int}_t,y^{int}_t)},
\end{equation}
where $t_s$ and $t_e$ represent the start and end time of the trajectory pair, respectively.

\subsection{Training Strategy}
\label{sec:training}
We adopt a four-step scheme for training.
\textbf{Step one:} We train the query-based Transformer decoder and prediction heads using video frames with frame-wise object and relationship annotations via the training loss \(\mathcal{L}_{step_1} = \mathcal{L}^o\), as detailed in Sec.~\ref{sec:loss1}. 
\textbf{Step two:} We train the auxiliary object classifier using video frames with provided ground-truth bounding boxes via the training loss \(\mathcal{L}_{step_2} = \mathcal{L}_{cls}^o\), as detailed in Sec.~\ref{sec:loss1}. 
\textbf{Step three:} We train the open-vocabulary relationship classifier using videos with provided ground-truth object trajectories via the training loss \(\mathcal{L}_{step_3} = \mathcal{L}^r\), as detailed in Sec.~\ref{sec:loss2}. 
\textbf{Step four:} We jointly fine-tune the entire end-to-end framework via the overall training loss
\(\mathcal{L}_{step_4} = \mathcal{L}^o + \mathcal{L}^r\).
{\subsection{Computational Complexity Analysis}}

The computational complexity of our framework is determined by three main stages: (1) frame-wise object detection, (2) trajectory association, and (3) spatio-temporal visual prompting for paired object trajectories. Below, we analyze each component in detail.

\noindent {\textbf{Frame-Wise Object Detection Complexity.} For each frame, objects are detected independently. Let $N_v$ be the number of frames in a video and $N_q$ be the number of object queries, which determines the maximum number of objects that the model can query in each frame, the computational complexity of this stage is $\mathcal{O}(N_v \cdot N_q)$. Since the input resolution is normalized to  a fixed size of $336 \times 336$ by CLIP's ViT encoder, the complexity is independent of the original frame resolution.}

\noindent {\textbf{Trajectory Association Complexity.} After objects are detected in each frame, trajectories are constructed by associating objects across frames. Let $N_o$ denote the number of objects detected per frame, which is typically much smaller than the number of object queries \( N_q \).  The trajectory association step involves pairwise comparisons of detected objects between consecutive frames, with a computational complexity of approximately $\mathcal{O}(N_v \cdot N_o^2)$.}

\noindent {\textbf{Spatio-Temporal Visual Prompting Complexity.} After trajectories are constructed, spatio-temporal visual prompting is performed on object pairs. Let $N_t$ be the total number of trajectories in the video,  and the maximum number of trajectory pairs is $N_t \times (N_t-1)$. The computational complexity of this step is about $\mathcal{O}(N_v^2 \cdot N_t^2)$, as temporal prompting requires modeling features across different frames, resulting in quadratic complexity with the number of frames \( N_v \).}

\noindent {\textbf{Overall Complexity.} Combining the above stages, the total computational complexity is $\mathcal{O}(N_v \cdot N_q + N_v \cdot N_o^2 + N_v^2 \cdot N_t^2)$.}

\noindent {\textbf{Scalability.} While frame-wise object detection scales linearly with \( N_v \) and \( N_q \), trajectory association grows quadratically with \( N_o \), and spatio-temporal visual prompting scales quadratically, especially with respect to \( N_v \) and \( N_t \). This poses scalability challenges for longer videos and larger object sets, which is a common challenge faced by current Open-VidVRD methods~\cite{gao2023compositional,yang2024multi}. In our future work, incorporating linear or group attention mechanisms can reduce the complexity of temporal modeling and improve efficiency in handling long videos. Additionally, designing a selection mechanism to identify the most likely trajectory pairs  for classification, rather than classifying relationships pairwise for all trajectories, can help improve the efficiency in scenarios with large object sets.}

\subsection{Discussion}
In this paper, we use CLIP for the Open-VidVRD task. Video-text pre-trained models such as InternVideo~\cite{wang2022internvideo} and VideoCLIP~\cite{xu2021videoclip} can also be applied to our framework.
Compared with them, CLIP trained with images and texts performs better in preserving critical details of object positions and appearances in video frames, enabling our framework to effectively capture subtle visual information for object trajectory detection, thereby facilitating relationship classification.

\section{Experiment}
\label{sec:experiment}
\subsection{Datasets and Evaluation Metrics}
\subsubsection{Datasets} 

We evaluate our method on the \textbf{VidVRD}~\cite{shang2017video} and \textbf{VidOR}~\cite{shang2019relation} datasets.
The VidVRD dataset contains 1000 videos, with 800 videos for training and 200 for testing, covering 35 object categories and 132 predicate categories. The average video length in VidVRD is 9.7 seconds. The VidOR dataset contains 10000 videos, with 7000 videos for training, 835 for validation, and 2165 for testing, covering 80 object categories and 50 predicate categories. The videos in VidOR are much longer, with an average length of 34.6 seconds.

\subsubsection{Evaluation Settings}
For the open-vocabulary evaluation, the base and novel categories are selected based on frequency. Following RePro~\cite{gao2023compositional}, we choose the common object and relationship categories as base categories and the rare ones as novel categories. 
Training is performed on the base categories and testing is performed under two settings: (1) \textbf{Novel}-split evaluation involves novel object categories for trajectory detection, and all object categories along with novel relationship categories for relationship classification. (2) \textbf{All}-split evaluation involves all object categories and all relationship categories, which is a standard evaluation. Note that the test is performed on both the VidVRD test set and the VidOR validation set (the annotations of the VidOR test set are not available). 
\subsubsection{Evaluation Tasks} 
Following Motif-Net~\cite{zellers2018neural}, we evaluate the model on three standard VidVRD tasks: scene graph detection (\textbf{SGDet}), scene graph classification (\textbf{SGCls}), and predicate classification (\textbf{PredCls}). Specifically, SGDet detects object trajectories from raw videos and classifies the relationships between these objects.
SGCls classifies the objects within the provided ground-truth trajectories and then predicts the relationships between these objects.
PredCls predicts the relationships between known objects, where both the ground-truth trajectories and corresponding object categories are provided. 

\subsubsection{Metrics} 
We use mean Average Precision (\textbf{mAP}) and Recall@K (\textbf{R@K}) with K = 50, 100 as evaluation metrics for relationship classification. The detected triplet is considered correct if it matches a ground-truth triplet and the IoU between the trajectories is greater than a threshold (\textit{i.e.,} 0.5). These metrics are applied across all tasks. For SGDet and SGCls tasks, we introduce an additional metric, called mean Average Precision of object trajectory (\textbf{mAP$_o$}), to evaluate the quality of object trajectories.

\subsection{Implementation Details}
For all experiments, video frames are sampled every 30 frames. We adopt the ViT-L/14 version of CLIP with fixed parameters. 

For the query-based Transformer decoder, we use six Transformer layers and 300 object queries. The temperature parameter \(\tau\) in both Eq.~\ref{eq:mc} and Eq.~\ref{eq:mc2} is set to 0.01. The threshold \(\epsilon\) used to filter the bounding boxes is set to 0.35.
For the auxiliary object classifier, we use eight tokens each for learnable continuous prompts and learnable conditional prompts, positioning the object token [OBJ] at the end of the sequence.
The weighting factors $\alpha$ and $\beta$ in Eq.~\ref{eq:weighting1} are set to 0.3 and 0.6, respectively.
The coefficient \(\lambda_s\) in Eq.~\ref{eq:weighting3} is set to two. 
The coefficients $\lambda_1$, $\lambda_2$, $\lambda_3$, $\lambda_4$ and $\lambda_5$ in Eq.~\ref{eq:weighting2} are set to three, five, five, two and two, respectively.
The number of Transformer blocks for spatio-temporal visual prompting is set to one for VidVRD and two for VidOR. 
For the open-vocabulary relationship classifier, we use eight tokens each for learnable continuous prompts and learnable conditional prompts, positioning the relationship token [REL] at 75\% of the token length. 
The coefficients \(\gamma\) and \(\delta\) in Eq.~\ref{eq:weighting4} are set to 0.2 and 0.1, respectively. 

For all training steps in Sec.~\ref{sec:training}, we use the AdamW~\cite{loshchilov2017decoupled} algorithm for optimization. 
In step one, we initialize the query-based Transformer decoder, object embedding head, and bounding box regression head with pre-trained parameters from MS-COCO~\cite{lin2014microsoft} (excluding novel object categories in Open-VidVRD), while the relationship classification head is initialized with random parameters. The learning rate is set to 1e-5, and the model is trained for ten epochs with a batch size of 16.
In step two, the auxiliary object classifier is trained for five epochs 
 with a learning rate of 1e-3 and a batch size of 12.
In step three, the open-vocabulary relationship classifier is trained with an initial learning rate of 1e-4, following a multi-step decay schedule that reduces the learning rate by a factor of 0.1 at epochs 15, 20, and 25, with a batch size of 32.
In step four, we fine-tune the end-to-end framework for 5 epochs with an initial learning rate of 1e-5 and a batch size of one.

\begin{table*}
\caption{{Results of different methods on the VidVRD dataset. }}
\centering
\begin{tabular}{c|c|cccc|cccc|ccc} 
\hline \hline
\multirow{2}{*}{Split} & \multirow{2}{*}{Method} & \multicolumn{4}{c|}{SGDet} & \multicolumn{4}{c|}{SGCls}& \multicolumn{3}{c}{PredCls}\\
\cline{3-13}
\multicolumn{1}{c|}{}&\multicolumn{1}{c|}{}&mAP$_o$&mAP&R@50&R@100&mAP$_o$&mAP&R@50&R@100&mAP&R@50&R@100\\
\hline
\multirow{6}{*}{Novel}
& ALPro
&10.36&0.98&2.79&4.33&21.06&3.69&7.27&8.92&4.09&9.42&10.41\\
& {CLIP}
&{14.37}&{2.13}&{3.26}&{4.50}&{24.96}&{3.84}&{6.03}&{9.44}&{4.54}&{7.27}&{11.74}\\
&VidVRD-II
&10.36&3.11&7.93&11.38&21.06&5.70&13.22&18.34&7.35&18.84&26.44\\
&RePro
&10.36&5.87&12.75&16.23&21.06&10.32&19.17&25.28&12.74&25.12&33.88\\
&OV-MMP
&14.37&12.15&13.72&15.21&24.96&17.57&21.98&28.43&21.14&30.41&37.85\\
&Ours&\textbf{36.31}&\textbf{15.04}&\textbf{16.03}&\textbf{18.18}&\textbf{31.73}&\textbf{17.96}&\textbf{30.74}&\textbf{36.86}&\textbf{21.65}&\textbf{35.37}&\textbf{43.64}\\
\hline
\hline
\multirow{6}*{All}
&ALPro
&18.18&3.03&2.57&3.11&68.99&3.92&3.88&4.75&4.97&4.50&5.79\\
& {CLIP}
&{34.61}&{4.86}&{2.97}&{3.55}&{70.39}&{5.80}&{4.37}&{5.38}&{6.49}&{5.21}&{6.54}\\
&VidVRD-II
&18.18&12.66&9.72&12.50&68.99&17.26&14.93&19.68&19.73&18.17&24.90\\
&RePro
&18.18&21.12&12.63&15.42&68.99&30.15&19.75&25.00&34.90&25.50&32.49\\
&OV-MMP
&34.61&22.10&13.26&16.08&70.39&29.38&23.56&28.89&38.08&30.47&37.46\\
&Ours&\textbf{52.72}&\textbf{26.34}&\textbf{16.48}&\textbf{19.54}&\textbf{74.25}&\textbf{31.95}&\textbf{25.96}&\textbf{31.66}&\textbf{39.83}&\textbf{31.66}&\textbf{39.69}\\
\hline \hline
\end{tabular}

\label{tab:sota-vidvrd}
\end{table*}

\begin{table*}
\caption{{Results of different methods on the VidOR dataset. For ALPro, VidVRD-II, and RePro, only the results of R@50 and R@100 on the \textbf{SGCls} and \textbf{PredCls} tasks are available from their original papers.}}
\centering
\begin{tabular}{c|c|cccc|cccc|ccc} 
\hline \hline
\multirow{2}{*}{Split} & \multirow{2}{*}{Method} & \multicolumn{4}{c|}{SGDet} & \multicolumn{4}{c|}{SGCls}& \multicolumn{3}{c}{PredCls}\\
\cline{3-13}
\multicolumn{1}{c|}{}&\multicolumn{1}{c|}{}&mAP$_o$&mAP&R@50&R@100&mAP$_o$&mAP&R@50&R@100&mAP&R@50&R@100\\
\hline
\multirow{6}{*}{Novel}
& ALPro
&-&-&-&-&-&-&3.17&3.74&-&8.35&9.79\\
& {CLIP}
&{1.11}&{0.17}&{0.68}&{0.77}&{6.04}&{0.43}&{1.79}&{2.36}&{1.08}&{5.48}&{7.20}\\
&VidVRD-II
&-&-&-&-&-&-&1.44&2.01&-&4.32&4.89\\
&RePro
&-&-&-&-&-&-&2.01&2.30&-&7.20&8.35\\
&OV-MMP
&1.11&0.84&1.44&1.44&6.04&2.40&5.48&6.92&3.58&9.22&11.53\\
&Ours&\textbf{2.33}&\textbf{2.45}&\textbf{4.79}&\textbf{4.79}&\textbf{6.83}&\textbf{2.72}&\textbf{5.76}&\textbf{8.65}&\textbf{4.11}&\textbf{9.80}&\textbf{14.41}\\
\hline
\hline
\multirow{6}*{All}
&ALPro
&-&-&-&-&-&-&0.95&1.32&-&2.61&3.66\\
& {CLIP}
&{3.38}&{0.22}&{0.35}&{0.51}&{25.86}&{0.63}&{0.74}&{0.99}&{1.29}&{1.71}&{3.13}\\
&VidVRD-II
&-&-&-&-&-&-&9.40&12.78&-&24.81&34.11\\
&RePro
&-&-&-&-&-&-&10.03&12.91&-&27.11&35.76\\
&OV-MMP
&3.38&7.15&6.54&8.29&25.86&24.00&23.04&30.14&38.52&33.44&43.80\\
&Ours&\textbf{12.99}&\textbf{11.08}&\textbf{8.43}&\textbf{9.82}&\textbf{26.17}&\textbf{25.21}&\textbf{23.78}&\textbf{30.17}&\textbf{39.75}&\textbf{33.68}&\textbf{43.87}\\
\hline \hline
\end{tabular}
\label{tab:sota-vidor}
\end{table*}

\subsection{Comparison with Existing Methods}
We compare our method with existing  Open-VidVRD methods, including RePro~\cite{gao2023compositional}, VidVRD-II~\cite{shang2021video}, \textcolor{blue}{CLIP}~\cite{radford2021learning}, and ALPro~\cite{li2022align}, and our previous work OV-MMP~\cite{yang2024multi}.
Existing methods rely on trajectory detectors pre-trained on closed datasets that encompass all object categories in Open-VidVRD. To ensure a fair comparison, we reproduce the compared existing methods by removing the novel object categories from the training data and retraining the trajectory detector.
On the VidOR dataset, the models and codes of ALPro, VidVRD-II, and RePro have not been ready to use, and only the results for R@50 and R@100 are available from their original papers.

Table~\ref{tab:sota-vidvrd} and Table~\ref{tab:sota-vidor} show the comparison results on the VidVRD and VidOR datasets, respectively. 
We have several interesting observations as follows: (1) Our method  outperforms all existing methods that rely on trajectory detectors pre-trained on closed datasets across all metrics on both datasets, demonstrating the superiority of  unifying object trajectory detection and relationship classification in an end-to-end framework for Open-VidVRD;  
(2) For the novel split, our method consistently achieves the best results across all datasets, especially improving mAP$_o$ by 21.94\% and mAP by 2.89\%  on the SGDet task on the VidVRD dataset, and improving mAP$_o$ by  1.22\% and mAP by 1.61\% on the VidOR dataset.  This highlights its strong generalization capability in open-vocabulary scenarios, which benefits from the proposed relationship-aware open-vocabulary trajectory detector and the proposed multi-modal prompting based open-vocabulary relationship classifier; 
On both SGCls and PredCls tasks where the ground-truth object trajectories are provided, our method also achieves better performance than the existing methods, which suggests that the open-vocabulary relationship classifier benefits from joint learning of trajectory detection and relationship classification.
(4) On the VidVRD dataset, for the novel split, our method achieves higher  mAP$_o$  on the SGDet task than on the SGCls task. This is because the detected bounding boxes are of high quality, and the classification results of SGDet benefit from ensembling the classification results of both the object queries and the auxiliary object classifier, whereas the results of SGCls rely solely on the auxiliary object classifier.
In contrast, for the VidOR dataset, the mAP$_o$ results on the SGDet task are limited by the trajectory association due to more blurs and occlusions in longer videos.

\begin{table}[t]
\caption{Performance (mAP$_o$ and mAP) of ablation study for end-to-end training on the VidVRD and VidOR datasets.}
\centering
\begin{tabular}{c|c|cc|cc}
\hline
\hline
\multirow{2}{*}{Dataset} &\multirow{2}{*}{\shortstack{End-to-end\\training}} &  \multicolumn{2}{c|}{Novel} & \multicolumn{2}{c}{All}\\
\cline{3-6}
&&mAP$_o$&mAP&mAP$_o$&mAP\\
\hline
\multirow{2}{*}{VidVRD}&&33.06&14.43&46.76&25.39\\
&\checkmark&\textbf{36.31}&\textbf{15.04}&\textbf{52.72}&\textbf{26.34}\\
\hline
\multirow{2}{*}{VidOR}&&1.00&1.99&10.14&10.45\\
&\checkmark&\textbf{2.33}&\textbf{2.45}&\textbf{12.99}&\textbf{11.08}\\
\hline
\hline
\end{tabular}
\label{tab:end2end}
\end{table}

\subsection{Ablation Studies}
\subsubsection{Effectiveness of End-to-end Training}
To evaluate the effectiveness of the end-to-end training of the trajectory detector and relationship classifier, we design a separate training strategy for comparison, where the trajectory detector is trained using the frame-wise object and relationship annotations, and the relationship classifier is trained using the ground-truth trajectories and video-level relationship annotations. Table~\ref{tab:end2end} shows the results on the VidVRD and VidOR datasets, and the proposed end-to-end training performs better in both trajectory detection and relationship classification, further verifying the advantage of unifying trajectory detection and relationship classification. 

\begin{table}[t]
\caption{Performance (mAP$_o$ and mAP) of ablation study for the relationship-aware open-vocabulary trajectory detector. ``Rna" denotes the relationship query and corresponding auxiliary relationship loss. ``Aoc" denotes the auxiliary object classifier.}
\centering
\begin{tabular}{cc|cc|cc}
\hline
\hline
\multirow{2}{*}{Rna} & \multirow{2}{*}{Aoc}& \multicolumn{2}{c|}{Novel} & \multicolumn{2}{c}{All}\\
\cline{3-6}
 & &mAP$_o$&mAP&mAP$_o$&mAP\\
\hline
 & &26.17&14.23&43.09&24.57\\
\checkmark& &27.33&14.56&46.16&25.36\\
 &\checkmark&30.29&14.58&48.11&25.47\\
\checkmark &\checkmark&\textbf{36.31}&\textbf{15.04}&\textbf{52.72}&\textbf{26.34}\\
\hline
\hline
\end{tabular}
\label{tab:abl_obj_enh}
\end{table}
\begin{table}[t]
\caption{Performance (mAP) of ablation study for multi-modal prompting on the VidVRD dataset. ``Vis" and ``Lan" denote visual prompting and language prompting, respectively.}
\centering
\begin{tabular}{cc|cc|cc}
\hline
\hline
\multirow{2}{*}{Vis} & \multirow{2}{*}{Lan}& \multicolumn{2}{c|}{Novel} & \multicolumn{2}{c}{All}\\
\cline{3-6}
 & &SGDet&PredCls&SGDet&PredCls\\
\hline
 & &9.14&13.36&22.86&35.24\\
\checkmark& &11.46&14.59&24.28&36.64\\
 &\checkmark&9.72&15.60&24.81&38.38\\
\checkmark &\checkmark&\textbf{15.04}&\textbf{21.65}&\textbf{26.34}&\textbf{39.83}\\
\hline
\hline
\end{tabular}
\label{tab:abl_multimodal}
\end{table}

\begin{table}[t]
\caption{Performance (mAP) of ablation study for the spatio-temporal visual prompting on the VidVRD dataset. 
``Spa" and ``Tem" denote spatial modeling and temporal modeling, respectively.}
\centering
\begin{tabular}{cc|cc|cc}
\hline
\hline
\multirow{2}{*}{Spa} & \multirow{2}{*}{Tem}& \multicolumn{2}{c|}{Novel} & \multicolumn{2}{c}{All}\\
\cline{3-6}
 & &SGDet&PredCls&SGDet&PredCls\\
\hline
 & &9.72&15.60&24.81&38.38\\
\checkmark& &12.14&17.91&25.02&37.22\\
 &\checkmark&11.58&16.59&23.47&34.32\\
\checkmark &\checkmark&\textbf{15.04}&\textbf{21.65}&\textbf{26.34}&\textbf{39.83}\\
\hline
\hline
\end{tabular}

\label{tab:abl_visual}
\end{table}

\subsubsection{Effectiveness of Relationship-aware Open-vocabulary Trajectory Detector}
We propose a relationship query and a corresponding auxiliary relationship loss (denoted as ``Rna") to help the trajectory detector explicitly perceive the relationships between objects. We further use an auxiliary object classifier (denoted as ``Aoc") to enhance the object classification of the trajectory detector. 
The ablation study results of ``Rna" and ``Aoc" on the VidVRD dataset are shown in Table~\ref{tab:abl_obj_enh}. It is interesting to observe that both the proposed relationship query with the corresponding loss and the introduced object classifier enhance the relationship detection performance.

\subsubsection{Effectiveness of Multi-modal Prompting}
To evaluate the multi-modal prompting, we replace the spatio-temporal visual prompting (denoted as ``Vis") with linear layers and replace the vision-guided language prompting (denoted as ``Lan") with handcraft language prompting for comparison. The results on the VidVRD dataset are reported in Table~\ref{tab:abl_multimodal}, demonstrating the effectiveness of the proposed visual promoting and language prompting. 

\subsubsection{Effectiveness of Spatio-temporal Visual Prompting}
To further evaluate the spatio-temporal visual prompting, we replace the spatial modeling module (denoted as ``Spa") or the temporal modeling module (denoted as ``Tem")  with linear layers. 
According to the results presented in Table~\ref{tab:abl_visual}, our method achieves a 2.9\% improvement in mAP on the novel split for the SGDet task when performing both spatial and temporal modeling. Furthermore, we observe that the performance drops significantly when only temporal modeling is performed without incorporating spatial modeling. This is in line with expectations, as it is difficult to recognize object relationships based only on the dynamic state changes of individual objects.

\begin{table}[t]
\caption{Performance (mAP$_o$ and mAP) of ablation study for the vision-guided language prompting on the VidVRD dataset.}
\centering
\begin{tabular}{c|cc|cc}
\hline
\hline
\multirow{2}{*}{Variants} &  \multicolumn{2}{c|}{Novel} & \multicolumn{2}{c}{All}\\
\cline{2-5}
&mAP$_o$&mAP&mAP$_o$&mAP\\
\hline
Manual&31.97&11.46&49.69&24.28\\
Continuous&35.66&12.79&51.86&25.56\\
Conditional &34.83&13.40&50.96&25.52\\
Ours &\textbf{36.31}&\textbf{15.04}&\textbf{52.72}&\textbf{26.34}\\
\hline
\hline
\end{tabular}
\label{tab:abl_language_obj}
\end{table}

\begin{figure}[t]
\centering
\includegraphics[width=0.95\linewidth]{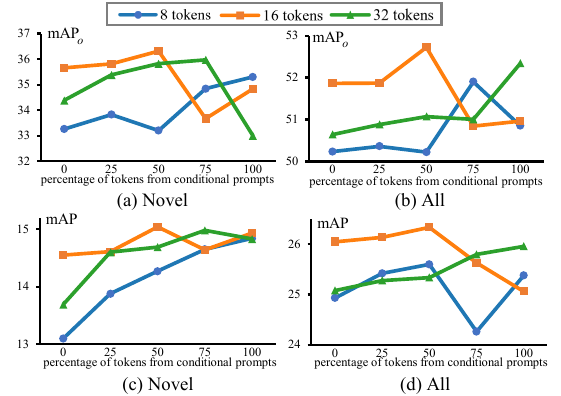}
\caption{Results of different token numbers of vision-guided language prompts on the VidVRD dataset. Different colors denote different token numbers, \textit{i.e.}, the blue, orange, and green colors represent the 8,16, and 32 tokens. The horizontal axis represents the percentage of tokens from conditional prompts, \textit{i.e.}, from 0 (all tokens are from learnable continuous prompts) to 100\% (all tokens are from learnable conditional prompts).  (a) and (b) show the results of using different tokens on the mAP$_o$ metric in the auxiliary object classifier. (c) and (d) show the results of using different tokens on the mAP metric in the open-vocabulary relationship classifier.}
\label{fig:abl_token}
\end{figure}

\begin{table}[t]
\caption{Parameter analysis results (mAP$_o$ and mAP) of $\epsilon$ on the VidVRD dataset.}
\centering
\begin{tabular}{c|cc|cc}
\hline
\hline
\multirow{2}{*}{$\epsilon$} &  \multicolumn{2}{c|}{Novel} & \multicolumn{2}{c}{All}\\
\cline{2-5}
&mAP$_o$&mAP&mAP$_o$&mAP\\
\hline
0.20&33.57&14.18&51.93&25.91\\
0.25&34.81&14.93&52.40&26.19\\
0.30 &35.15&14.89&52.69&26.30\\
0.35 &\textbf{36.31}&\textbf{15.04}&\textbf{52.72}&\textbf{26.34}\\
0.40&34.86&14.61&50.27&25.98\\
0.45&32.24&12.95&45.90&20.89\\
\hline
\hline
\end{tabular}
\label{tab:epo}
\end{table}

\subsubsection{Effectiveness of Vision-guided Language Prompting}
To further evaluate the  vision-guided language prompting, we design three variants of our method for comparison: (1) ``Manual" involves pre-defined templates for the auxiliary object classifier (\textit{i.e.}, ``An image of a [OBJ]") and the relationship classifier (\textit{i.e.}, ``An image of a person or object [REL] something" for subjects, ``An image of something [REL] a person or object” for objects, and ``An image of the visual relationship [REL] between two objects" for background); (2) ``Continuous" involves learnable continuous prompts; (3) ``Conditional" tailors all prompts to input visual features. 
The results in Table~\ref{tab:abl_language_obj} demonstrate that integrating the proposed vision-guided language prompting (``Ours") into the auxiliary object classifier and the relationship classifier significantly enhances the performances of object trajectory classification and relationship classification.
Notably, there is an improvement of over 1.6\% in mAP in the novel split when using detected trajectories.

\begin{figure}[t]
\centering
\includegraphics[width=0.8\linewidth]{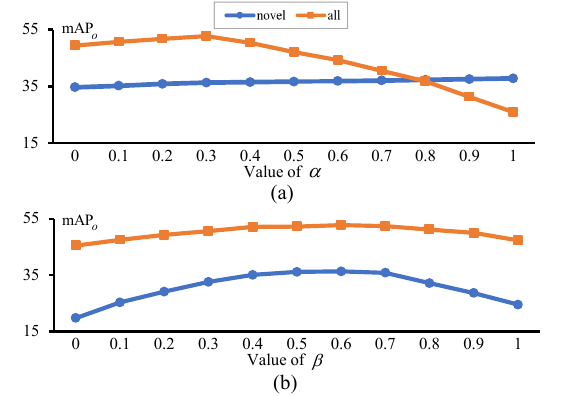}
\caption{Results of different values of the hyperparameters $\alpha$ and $\beta$ on the VidVRD dataset. The horizontal axis represents the values of the parameter, and the vertical axis represents the mAP$_o$  performance. (a) shows the results of different values of $\alpha$ while keeping $\beta=0.5$. (b) shows the results of different values of $\beta$ while keeping $\beta=0.3$.}
\label{fig:alpha}
\end{figure}

\subsection{Hyperparameters}
\subsubsection{The Token Number of Vision-guided Language Prompts}
To analyze the effects of different token numbers of vision-guided language prompts on performance, we conduct experiments using 8, 16, and 32 tokens for comparison. 
We also set the percentage of tokens from the learnable conditional prompts to 0, 25\%, 50\%, 75\%, and 100\% for comparison. 
Figure~\ref{fig:abl_token} shows the results of mAP$_o$ and mAP on the VidVRD dataset. 
We observe that as the number of tokens increases, the performance first increases and then decreases, with the best performance when the number of tokens is 16. 
We also observe that as the percentage of tokens from learnable conditional prompts increases,  the results first increase and then become unstable, and the result is best when half of the tokens come from learnable conditional prompts. 
These observations highlight the importance of combining task-specific knowledge and visual cues, further validating the effectiveness of the proposed vision-guided prompting in combining learnable continuous prompts and learnable conditional prompts.

\begin{figure*}[h]
\centering
\includegraphics[width=0.85\textwidth]{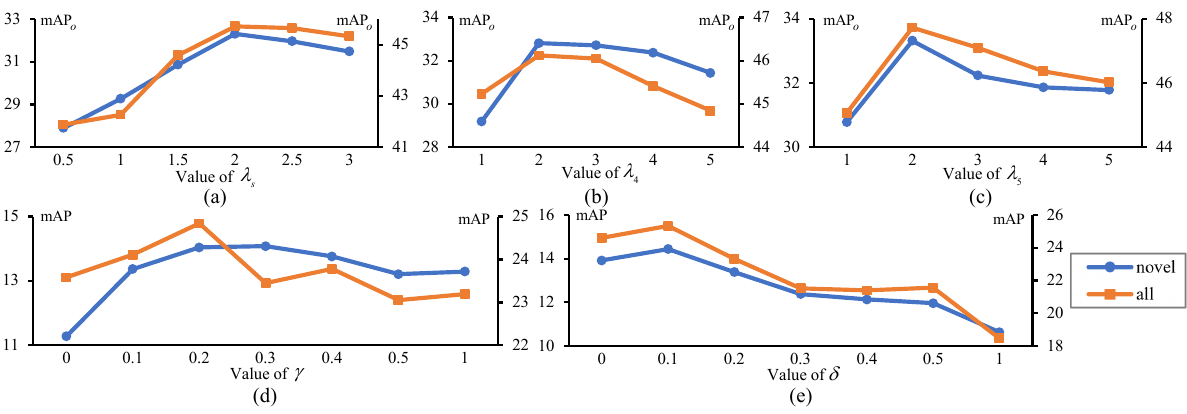}
\caption{
Results of different values of loss function coefficients on the VidVRD dataset. The horizontal axis represents the values of the coefficients. The left and right vertical axes represent the results of the  \textbf{novel} and \textbf{all}  categories.
(a) shows the results of different values of $\lambda_s$ while keeping $\lambda_4=1$ and $\lambda_5=1$. (b) shows the results of different values of $\lambda_4$ while keeping $\lambda_s=2$ and $\lambda_5=1$. (c) shows the results of different values of $\lambda_5$ while keeping $\lambda_4=2$ and $\lambda_5=2$. (d) shows the results of different values of $\gamma$ while keeping $\delta=0$. (e) shows the results of different values of $\delta$ while keeping $\gamma=0.2$.}
\label{fig:para}
\end{figure*}

\subsubsection{The Filtering Threshold of  Bounding Boxes }
To analyze the effect of the filtering threshold of bounding boxes in frame-wise open-vocabulary object detection, \textit{i.e.}, the hyperparameter $\epsilon$, we conduct experiments by varying the value of $\epsilon$ in $\{0.20,0.25,0.30,0.35,0.40,0.45\}$.  The results on the VidVRD dataset are shown in Table~\ref{tab:epo}. From these results, we observe that the performance initially improves as the threshold increases, but then decreases. This is because as $\epsilon$ increases, the exclusion of more false positive bounding boxes enhances performance. However, beyond a certain point, further increases in $\epsilon$ begin to eliminate true positive boxes, leading to a degradation in performance. The optimal value of $\epsilon$ is 0.35.

\subsubsection{The Coefficients for Ensembling Object Classification Results}
To analyze the effect of the coefficients for ensembling object classification results in frame-wise open-vocabulary object classficiation, \textit{i.e.}, the hyperparameters $\alpha$ and $\beta$ in Eq.~\ref{eq:weighting1}, we conduct experiments by varying the value of $\alpha$ in the range of $\{0.0,0.1,0.2,0.3,0.4,0.5,0.6,0.7,0.8,0.9,1.0\}$ while keeping $\beta=0.5$, and the results on the VidVRD dataset are shown in Figure~\ref{fig:alpha} (a). We observe that the optimal performance is achieved when $\alpha$ is set to 0.3. Then we vary the value of $\beta$ in the same range while keeping $\alpha = 0.3$, and the results are shown in Figure~\ref{fig:alpha} (b).
The overall performance reaches its peak value when $\beta$ is set to 0.6. It is worth noting that the ensemble is more effective for the {novel} categories, which can be seen from the significant impact of $\beta$ on the  mAP$_o$ results, highlighting the positive impact of the rich semantic information in CLIP on these categories.

\subsubsection{The Coefficients of Loss Functions}
According to DETR~\cite{carion2020end}, the coefficients \(\lambda_1\), \(\lambda_2\), and \(\lambda_3\) in Eq.~\ref{eq:weighting2} are set to three, five, and five, respectively. 
To analyze the impact of the other coefficients for loss functions of the relationship-aware open-vocabulary trajectory detector, \textit{i.e.}, \(\lambda_s\) in Eq.~\ref{eq:weighting3}, and \(\lambda_4\) and \(\lambda_5\) in Eq.~\ref{eq:weighting2}, we independently train the trajectory detector and vary \(\lambda_s\) in $\{0.5,1.0,1.5,2.0,2.5,3.0\}$, while keeping \(\lambda_4=1\) and \(\lambda_5=1\). The results on the VidVRD dataset are shown in Figure~\ref{fig:para} (a), indicating that the optimal performance is achieved when $\lambda_s$ is set to two. Then we vary $\lambda_4$ in $\{1,2,3,4,5\}$ with $\lambda_s = 2$ and $\lambda_5=1$, and the results are shown in Figure~\ref{fig:para} (b), which indicates  that the performance peaks when $\lambda_4$ is set to two. Subsequently, we vary $\lambda_5$ over $\{1,2,3,4,5\}$ with $\lambda_s = 2$ and $\lambda_4=2$. The results in Figure~\ref{fig:para} (c) show that the performance is maximized when
$\lambda_5$ is two.

Similarly, to analyze the effect of coefficients for loss functions of the open-vocabulary relationship classifier, \textit{i.e.}, the hyperparameters \(\gamma\) and \(\delta\) in Eq.~\ref{eq:weighting4}, we independently train the relationship classifier and vary $\gamma$ over $\{0.1,0.2,0.3,0.4,0.5,1.0\}$, while keeping $\delta=0$. The results on the VidVRD dataset are shown in Figure~\ref{fig:para} (d), which indicates that optimal performance is achieved when $\gamma$ is set to 0.2. Then we vary the value of $\delta$ in the same range while keeping $\gamma=0.2$, the results are shown in Figure~\ref{fig:para} (e). The best performance occurs when $\delta$ is set to 0.1.

\subsection{Qualitative Analysis}

\begin{figure*}[t]
	\centering
	\includegraphics[width=0.8\linewidth]{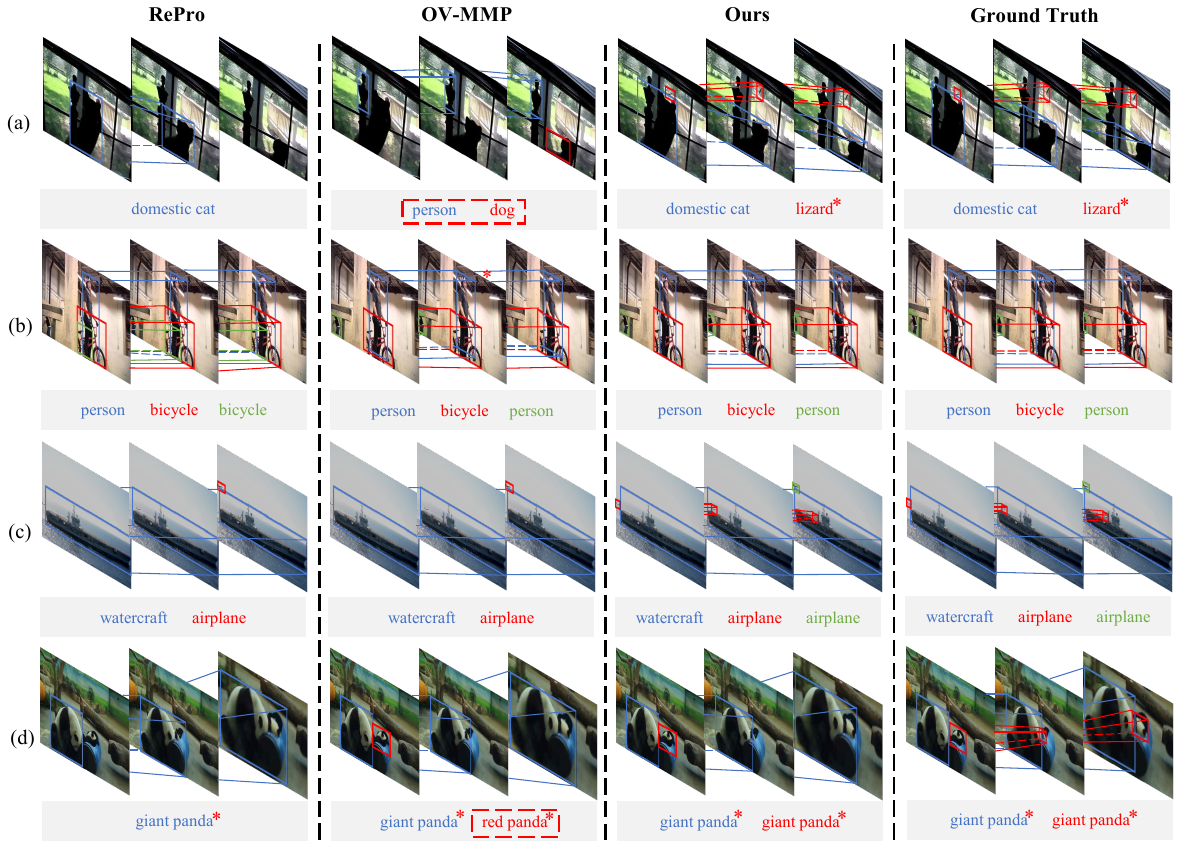}
	\caption{{Visualization of trajectories from different methods. The objects classified incorrectly are enclosed within the red dashed box. * represents  the novel object category.}}
	\label{fig:traj_vis}
\end{figure*}

\begin{figure}[t]
	\centering
	\includegraphics[width=0.9\linewidth]{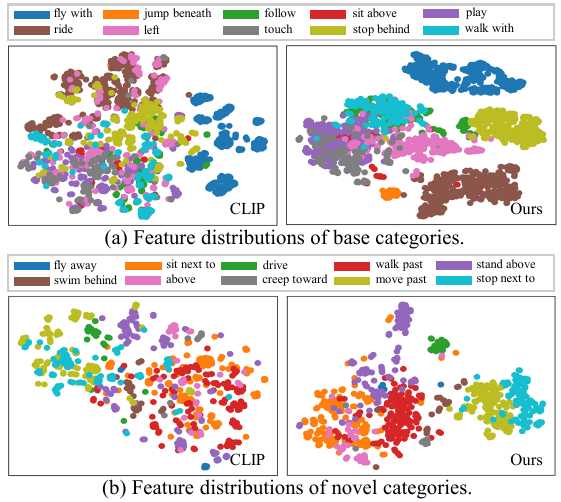}
	\caption{Qualitative results of visual feature (union region of subject and object) distributions by T-SNE.}
	\label{fig:feature}
\end{figure}
\subsubsection{Trajectory Visualization}
We visualize the trajectories generated by different methods on the VidVRD dataset. 
{Figure~\ref{fig:traj_vis} (a) shows a cat observing a lizard. RePro~\cite{gao2023compositional} and OV-MMP~\cite{yang2024multi} fail to detect the lizard, which belongs to a novel object category. Moreover, OV-MMP detects an incomplete trajectory of the cat and misidentifies an object out of interest, resulting in subsequent classification errors. In contrast, our method detects both objects accurately and classifies them correctly.
Figures~\ref{fig:traj_vis} (b) and (c) also show that both RePro and OV-MMP fail to detect certain objects, while our method detects all objects correctly without redundancy or omission.}
These examples demonstrate the strong generalization capability of our method to novel object categories and complex scenes.
{Figure~\ref{fig:traj_vis} (d) illustrates a case of severe occlusion, where none of the methods are able to fully detect the trajectory of the panda inside the barrel, suggesting the limitations of our method in such challenging scenarios. In the future, more advanced trajectory association algorithms could help improve trajectory detection performance by better handling occlusions and capturing motion dynamics over time.}

\subsubsection{Feature Distribution Visualization}
We visualize the feature distributions of randomly selected 10 predicate categories by projecting the features of the union regions onto a 2D plane using T-SNE~\cite{hinton2002stochastic}, to demonstrate how well our spatio-temporal visual prompting method adapts the image encoder of CLIP. 
As shown in Figure~\ref{fig:feature}, features of our method (the right parts of Figure~\ref{fig:feature} (a), (b)) within the same categories are pulled closer while features across different categories are pushed further apart, improving the discrimination on both base and novel categories.
These qualitative results further verify the effectiveness of our spatio-temporal visual prompting method. 
\subsubsection{{Relationship Visualization}}
\begin{figure*}
	\centering
	\includegraphics[width=0.85\linewidth]{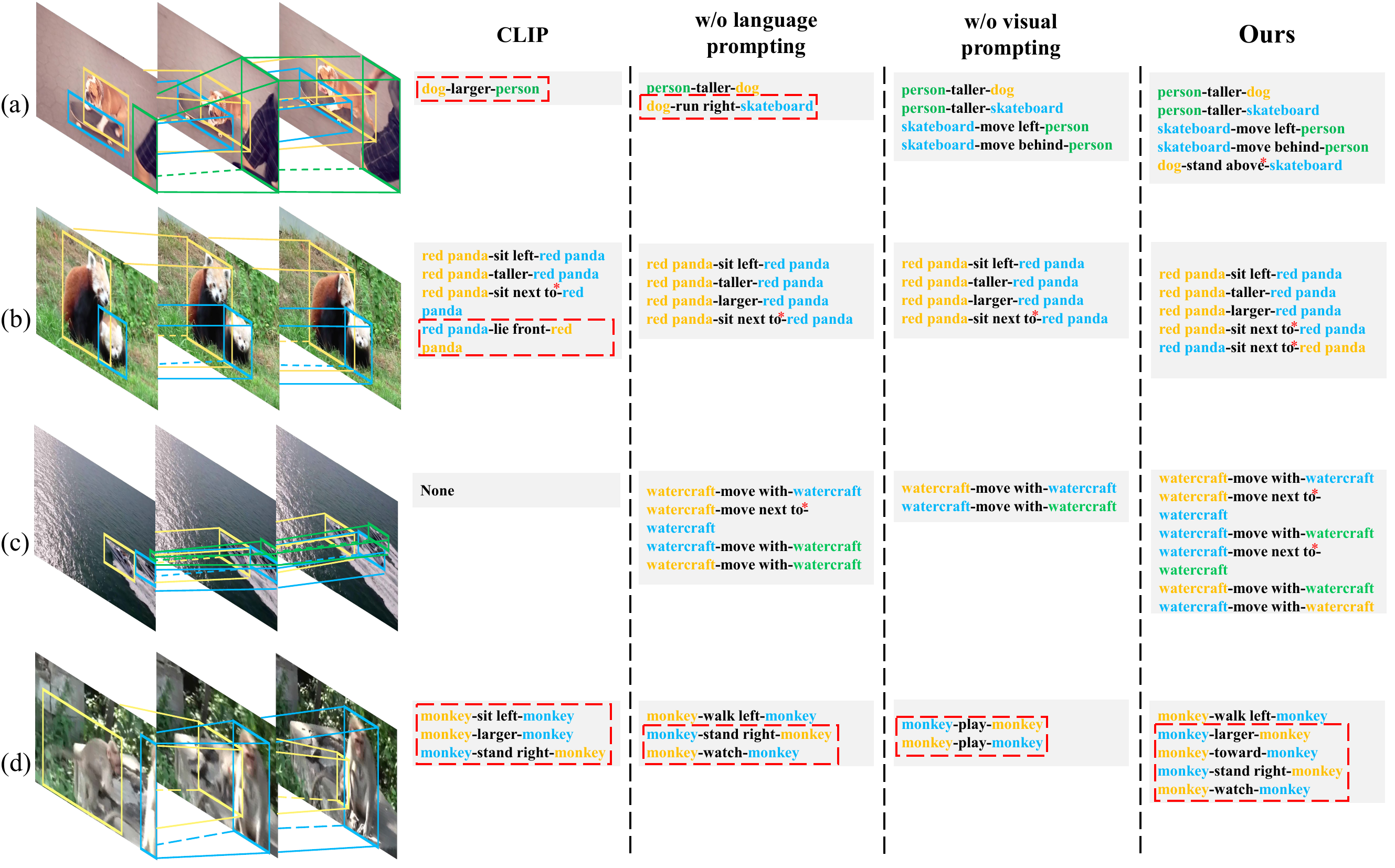}
	\caption{{Visualization of relationship classification results using ground truth trajectories. The relationships classified incorrectly are enclosed within the red dashed box. * represents  the novel relationship category.}}
	\label{fig:rel_case}
\end{figure*}
We visualize the correctly detected relationships using ground-truth trajectories on the VidVRD dataset. The comparison includes results from the original CLIP model, our method without language prompting (``w/o language prompting"), our method without visual prompting (``w/o language prompting"), and  our method.
Figure~\ref{fig:rel_case} (a), (b) and (c) show that CLIP performs poorly in relationship classification. Introducing language prompting or visual prompting can detect more correct relationships. Our method achieves the best performance, especially for the novel categories.
However, Figure~\ref{fig:rel_case} (d) shows a scene with severe occlusion and blur, where our method  only identifies one correct relationship. This suggests that there is still room for improvement, particularly in challenging scenarios with significant occlusion or ambiguity. Future advancements in multi-view fusion or more robust temporal modeling could enhance relationship classification accuracy.

\subsubsection{Case Studies}
\begin{figure*}[t]
	\centering
	\includegraphics[width=0.8\linewidth]{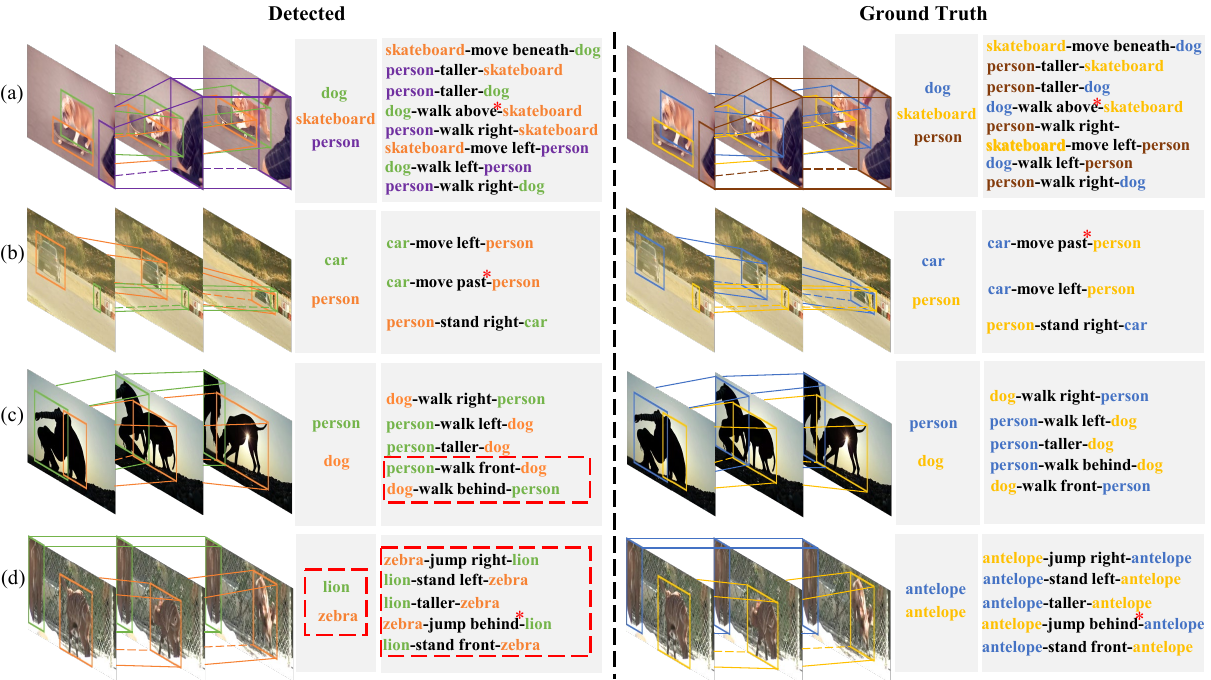}
	\caption{Examples of detected relationships by our method. The objects or relationship triplets detected incorrectly are enclosed within the red dashed box. * represents the novel relationship category.}
	\label{fig:case}
\end{figure*}
We conduct case studies on the VidVRD dataset to highlight the strengths of our end-to-end method and show the scenarios that lead to detection failures.
As shown in Figure~\ref{fig:case} (a) and (b), our method effectively detects object trajectories and accurately classifies the relationships between objects, including both base and novel categories, successfully overcoming challenges such as detecting partially visible objects and handling complex dynamic relationships like ``move past".
Figure~\ref{fig:case} (c) shows an example of video relationship detection in a backlit scene. Due to the strong sunlight, objects in the video appear as black silhouettes, losing visual texture information. Despite these challenging conditions, our method still successfully detects object trajectories, accurately classifies object categories, and correctly predicts most relationship categories, demonstrating its robustness to various scenes. However, due to the lack of texture information, it is difficult to determine the front-to-back positions of objects, resulting in some errors in relationship classification.
Figure~\ref{fig:case} (d) shows a challenging scene where the bodies of two antelopes are facing away from the camera, making object classification difficult. Our method incorrectly classifies the larger antelope on the left into a lion and the striped antelope on the right into a zebra. These misclassifications of objects lead to subsequent errors in triplet detection, despite the correct classification of the relationship categories.

\subsection{Cross-dataset Evaluation}
To evaluate the effectiveness of our method in detecting video relationships in real-world scenes, which exhibit a significant domain gap from the training set, we conduct a cross-dataset evaluation by training on the base categories of the VidOR dataset and testing directly on the VidVRD dataset, where categories that overlap with the training data are excluded.
In terms of categories, only 18 object categories in the VidVRD dataset appear in the novel categories of the VidOR dataset, while 17 object categories do not appear. Similarly, only 14 relationship categories in VidVRD appear in the novel categories of VidOR, while 118 relationship categories do not overlap. Additionally, the average video length differs significantly, with VidVRD videos averaging 9.7 seconds and VidOR videos averaging 34.6 seconds.
These significant discrepancies highlight the challenge of generalizing models to unseen object categories, unseen relationship categories, and unfamiliar video scenes, providing a rigorous evaluation of our method's performance in real-world video relationship detection scenarios.
\begin{table}
\centering
\caption{Comparison of cross-dataset transferred models and upper bound models on the SGDet task of the VidVRD dataset. }
\begin{tabular}{c|c|cc}
\hline
\hline
Setting&Method&mAP$_o$&mAP\\
\hline
\multirow{5}{*}{Cross dataset}&ALPro
&3.88&0.29\\
&VidVRD-II
&3.88&0.88\\
&RePro
&3.88&1.11\\
&OV-MMP
&2.74&1.14\\
&Ours&\textbf{13.65}&\textbf{6.59}\\
\hline
\multirow{5}{*}{Upper bound}&ALPro
&10.36&0.98\\
&VidVRD-II
&10.36&3.11\\
&RePro
&10.36&5.87\\
&OV-MMP
&14.37&12.15\\
&Ours&\textbf{36.31}&\textbf{15.04}\\

\hline
\hline
\end{tabular}
\label{tab:cross}
\end{table}

We present mAP$_o$ and mAP for object and relationship categories not seen in the training data as the results of cross-dataset experiments, using the results from the novel split without cross-dataset training as the upper bound, as shown in Table~\ref{tab:cross}. 
It can be observed that our method achieves the best results on all metrics in cross-dataset experiments, even surpassing the upper bound results of ALPro, VidVRD-II, and RePro, demonstrating the strong generalization capability of our end-to-end framework.

\section{Conclusion}
\label{sec:conclusion}
We present an end-to-end Open-VidVRD framework that unifies trajectory detection and relationship classification, eliminating the dependency on trajectory detectors pre-trained on closed datasets in the previous methods. 
Under this framework, we propose a relationship-aware open-vocabulary trajectory detector that can capture the relationship contexts via a relationship query and a corresponding auxiliary relationship loss to improve the trajectory detection performance. 
Moreover, we propose an open-vocabulary relationship classifier with a multi-modal prompting method that can prompt CLIP on both the visual and language sides to enhance the generalization to novel categories.
Experiments on VidVRD and VidOR datasets demonstrate significant improvements in the overall performance and generalization capability. 
In the future, we plan to unify trajectory detection and relationship classification within a Transformer decoder to further improve their mutual performance and make our method more practical for processing real-world videos, especially long videos with more complex object relationships.

\section*{Acknowledgments}
This work was supported in part by the National Natural Science Foundation of China (NSFC) under Grant 62072041.

\bibliographystyle{ieeetr}
\bibliography{references}

\begin{IEEEbiography}[{\includegraphics[width=1in,height=1.25in,clip,keepaspectratio]{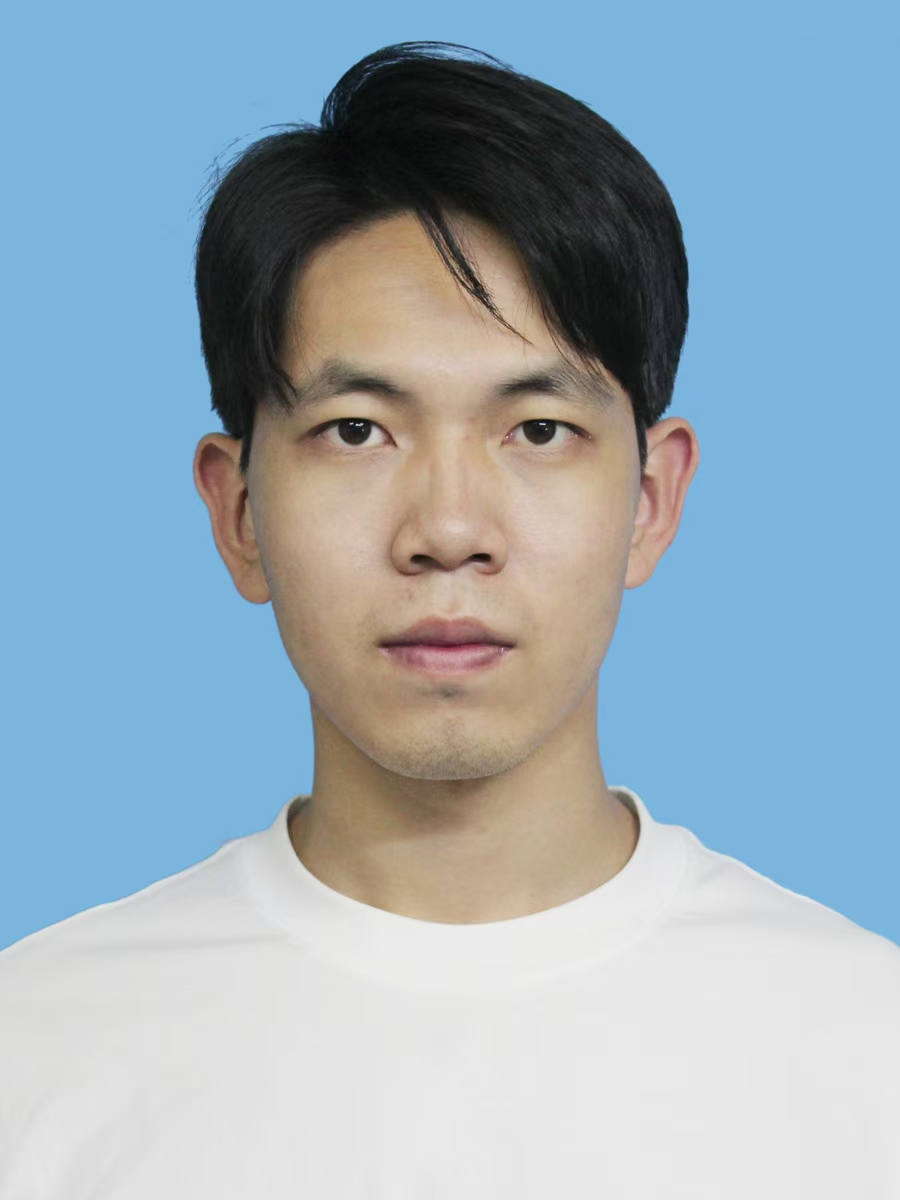}}]{Yongqi Wang}
received the B.S. degree in computer science in 2023 from the Beijing Institute of Technology (BIT),
Beijing, China, where he is currently working toward the M.S. degree in computer science. His research interests include vision and language, and video understanding.
\end{IEEEbiography}

\begin{IEEEbiography}[{\includegraphics[width=1in,height=1.25in,clip,keepaspectratio]{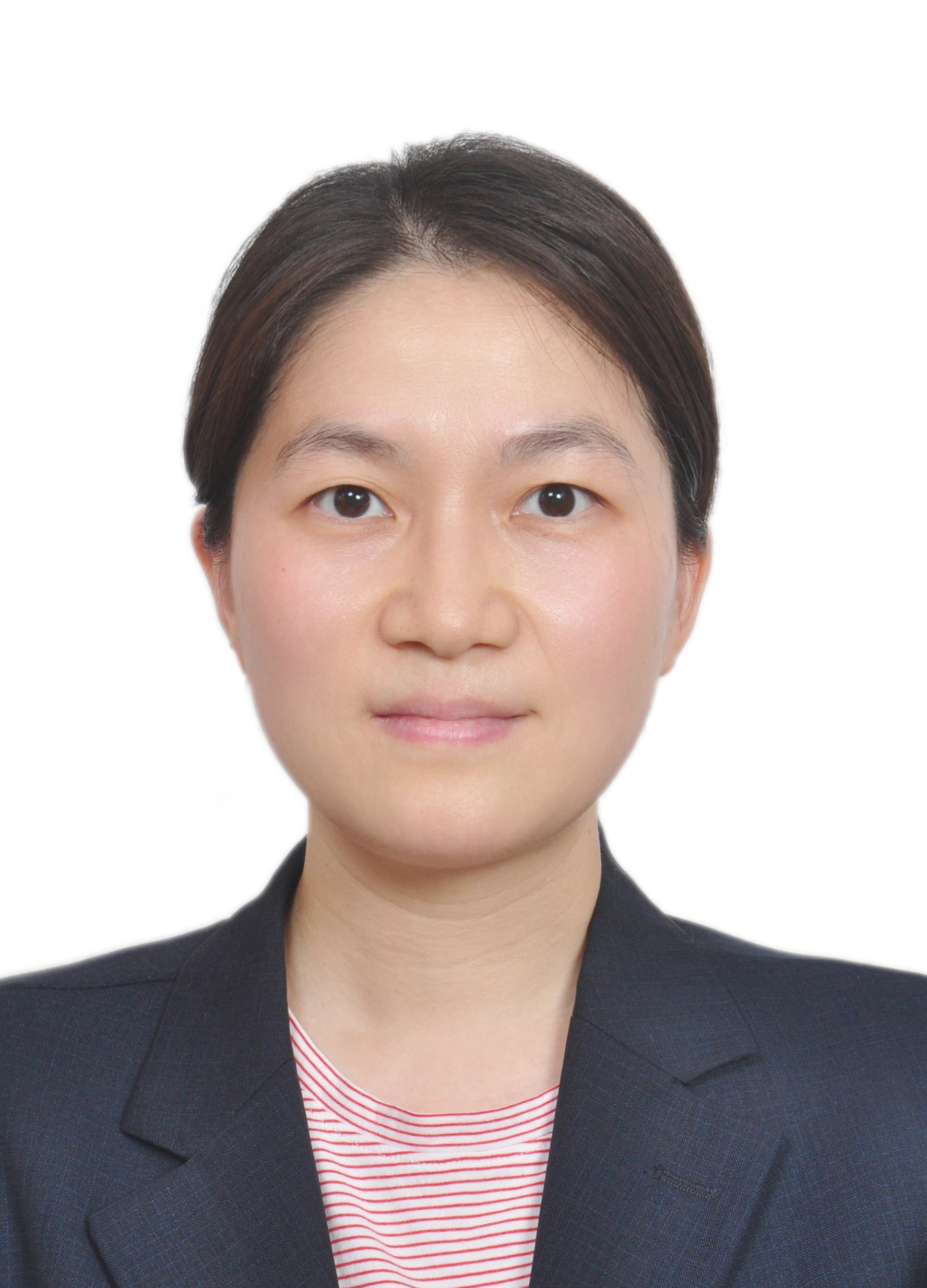}}]{Xinxiao Wu}
(Member, IEEE) received the B.S. degree in computer science from the Nanjing University of Information Science and Technology, Nanjing, China, in 2005, and the Ph.D. degree in computer science from the Beijing Institute of Technology (BIT), Beijing, China, in 2010. From 2010 to 2011, she was a Postdoctoral Research Fellow with Nanyang Technological University, Singapore. She is currently a Full Professor with the School of Computer Science, BIT. Her research interests include vision and language, machine learning, and video understanding. She serves on the Editorial Boards of the IEEE Transactions on Multimedia. 
\end{IEEEbiography}

\begin{IEEEbiography}[{\includegraphics[width=1in,height=1.25in,clip,keepaspectratio]{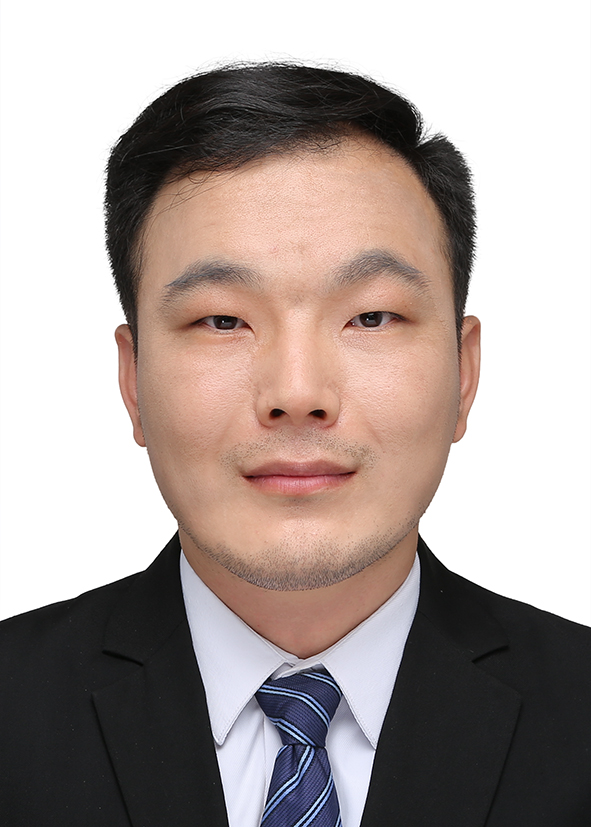}}]{Shuo Yang}
is an associate professor now at Shenzhen MSU-BIT University, Guangdong, China, from 2024. He received a B.S. degree in computer science from the Beijing Union University, Beijing, China, in 2014,  an M.S. degree in computer science from the Institute of Software, Chinese Academic of Science, Beijing, China, in 2017,  and a Ph.D. degree in computer science from the Beijing Institute of Technology (BIT),  Beijing, China, in 2024.  His research interests include visual understanding, and vision and language.
\end{IEEEbiography}

\begin{IEEEbiography}[{\includegraphics[width=1in,height=1.25in,clip,keepaspectratio]{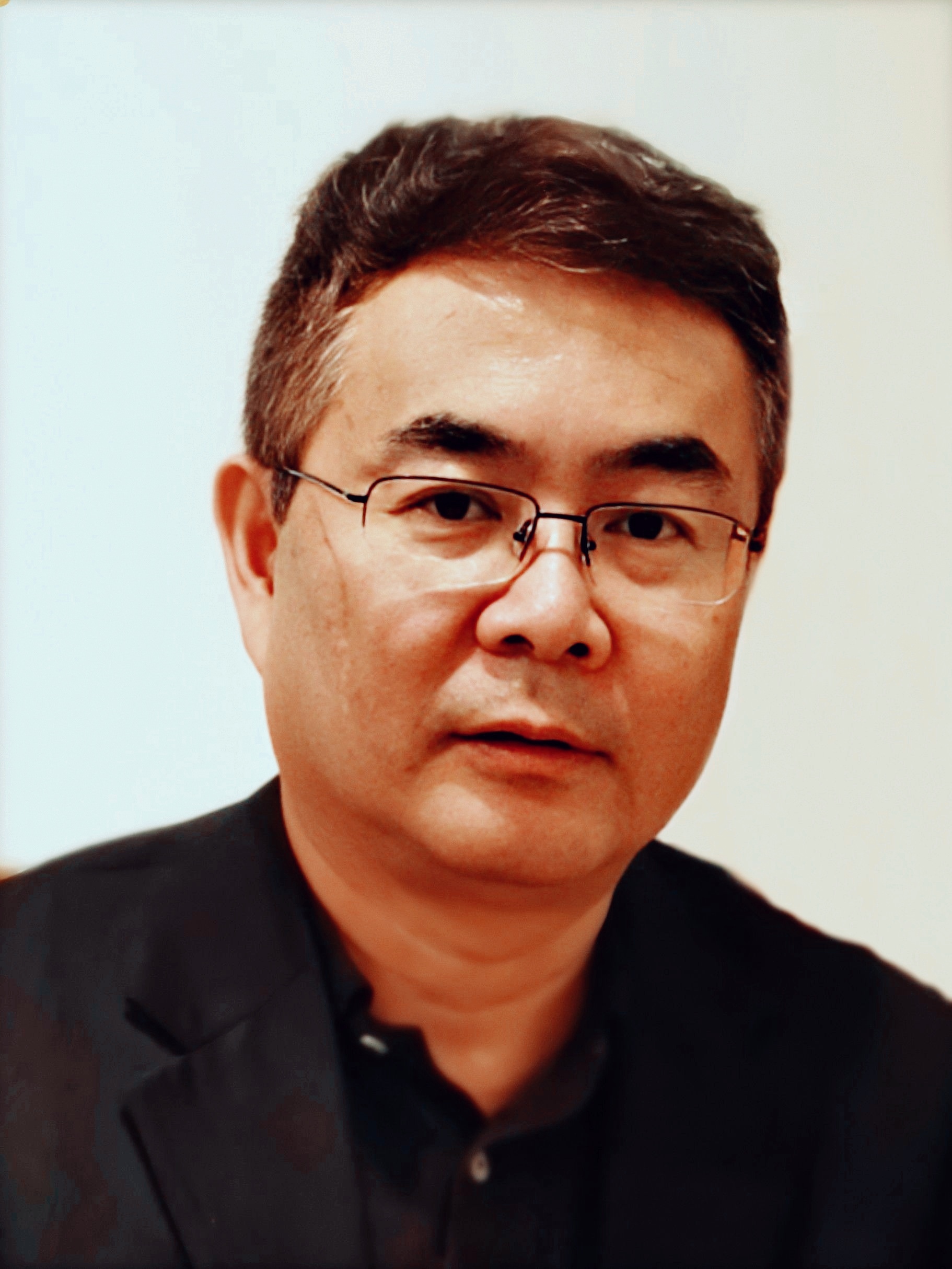}}]{Jiebo Luo}
(Fellow, IEEE) is currently the Albert Arendt Hopeman Professor of Engineering and Professor of Computer Science with the University of Rochester, Rochester, NY, USA, which he joined in 2011 after a prolific career of 15 years with the Kodak Research Laboratories. He has authored over 600 technical papers and holds more than 90 U.S. patents. His research interests include computer vision, natural language processing (NLP), machine learning, data mining, computational social science, and digital health. Dr. Luo is a fellow of NAI, ACM, AAAI, SPIE, and IAPR. He has been involved in numerous technical conferences, including serving as Program Co-Chair for ACM Multimedia 2010, IEEE CVPR 2012, ACM ICMR 2016, and IEEE ICIP 2017, and General Co-Chair for ACM Multimedia 2018 and IEEE ICME 2024. He has served on the editorial boards of IEEE TRANSACTIONS ON PATTERN ANALYSIS AND MACHINE INTELLIGENCE (TPAMI), IEEE TRANSACTIONS ON MULTIMEDIA (TMM), IEEE TRANSACTIONS ON CIRCUITS AND SYSTEMS FOR VIDEO TECHNOLOGY (TCSVT), IEEE TRANSACTIONS ON BIG DATA (TBD), ACM Transactions on Intelligent Systems and Technology (TIST), Pattern Recognition, Knowledge and Information Systems (KAIS), Machine Vision and Applications, and Intelligent Medicine. He was an Editor-in-Chief of IEEE TRANSACTIONS ON MULTIMEDIA from 2020 to 2022.
\end{IEEEbiography}

\end{document}